\documentclass{article} 
\usepackage{iclr2027_conference,times}

\usepackage{amsmath,amsfonts,bm}

\def\eqref#1{equation~\ref{#1}}

\def\1{\bm{1}}

\DeclareMathAlphabet{\mathsfit}{\encodingdefault}{\sfdefault}{m}{sl}
\SetMathAlphabet{\mathsfit}{bold}{\encodingdefault}{\sfdefault}{bx}{n}

\usepackage{hyperref}
\usepackage{url}

\title{A Broader Look at Model Merging: Rethinking Implicit Regularization Induced by Task Arithmetic}

\author{
\textbf{Sin-Han Yang}\textsuperscript{1}\thanks{Equal contribution}
\quad
\textbf{Shih-Cheng Huang}\textsuperscript{1}\footnotemark[1]
\quad
\textbf{Chieh-Yen Lin}\textsuperscript{1}
\quad
\textbf{Yun-Nung Chen}\textsuperscript{2}\thanks{Equal advisory contribution}
\\[0.6ex]
\hspace{0.1pt}
\textbf{Shao-Hua Sun}\textsuperscript{1,2}\footnotemark[2]
\quad
\textbf{Hung-yi Lee}\textsuperscript{2}\footnotemark[2]
\\[1.0ex]
\textsuperscript{1}Appier AI Research
\quad
\textsuperscript{2}National Taiwan University
\\
\texttt{sinhan.yang@appier.com, andy.huang@appier.com}
}

\iclrfinalcopy 

\usepackage[capitalize,noabbrev]{cleveref}
\usepackage{booktabs}
\usepackage{makecell}
\usepackage[table]{xcolor}
\usepackage{multirow}
\usepackage{amssymb}   
\usepackage{wrapfig}

\usepackage{etoc}

\usepackage[nottoc]{tocbibind}
\settocbibname{References}
\usepackage{minitoc}
\usepackage{bbm}

\usepackage{graphicx}
\usepackage{subcaption}

\begin{document}
\raggedbottom

\doparttoc
\faketableofcontents

\maketitle

\begin{abstract}

Model merging aims to build a multi-task model cheaply by combining the weights of individual task-specific models. To perform well across multiple tasks, most existing merging methods use an additional dataset to find the coefficients for the best linear combination of task-specific weight updates.
However, we identify an implicit regularization in this standard practice: searching over coefficients restricts the candidate models to a subspace spanned by task-specific weight updates. In this work, we investigate whether this regularization is actually useful. Surprisingly, empirical results show that in existing benchmarks, optimizing merged-model weights without this regularization significantly boosts the performance of common merging methods across multiple architectures and domains, even in an extremely data-limited scenario where only one instance is available per class. 
We further investigate strategies for using the additional dataset in different settings and discuss their practical use. We show that generalization is a key factor when choosing the regularization strength. 
Overall, this work calls for revisiting the existing model-merging pipeline, motivating a broader exploration of the weight space and a reconsideration of the implicit regularization induced by task arithmetic. Source code: \url{https://github.com/appier-research/broader_model_merging}.

\end{abstract}

\begin{figure}[h]
    \centering
    \includegraphics[width=\linewidth]{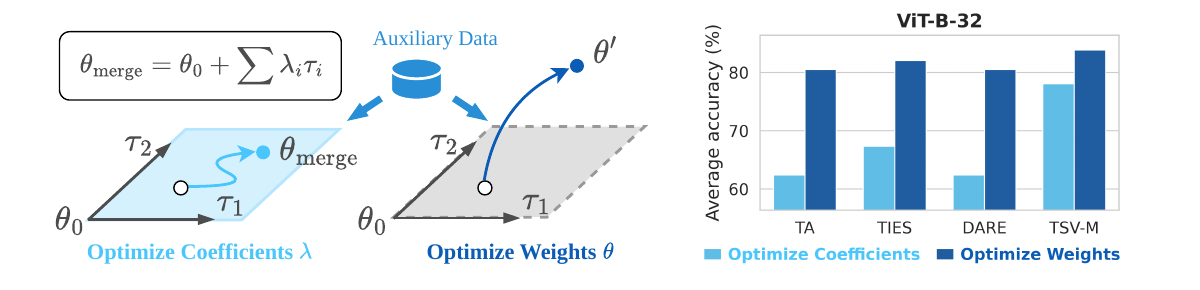}
    \caption{\textbf{A new perspective on the model-merging pipeline.} The existing model-merging pipeline builds the merged model $\theta_{merge}$ by adding a linear combination of task-specific weight updates $\tau_i$ to the pretrained model $\theta_0$. Standard practice uses an additional dataset, which we term \textit{auxiliary data}, to optimize the coefficients $\lambda$ for multi-task performance. This coefficient optimization regularizes model updates to the subspace spanned by $\tau_i$. \textbf{We challenge this pipeline by using the same auxiliary data to optimize the merged-model weights}, thereby removing the implicit regularization. In existing benchmarks, results show that exploring the broader weight space generally leads to significant performance improvements across merging methods (TA, TIES, DARE, and TSV-M).}
    \label{fig:fig1}
\end{figure}


\section{Introduction}

Model merging aims to build a unified model by combining the weights of individual task-specific models without retraining on full training data. It enables a wide range of applications, including multi-task model construction \citep{ilharco2023editing,yadav2023ties}, cross-lingual capability transfer \citep{huang2024chat,bandarkar2025layer}, and multi-objective preference alignment \citep{rame2023rewarded,jang2024personalized, liang2026adaptive}.

\textit{Task Arithmetic} \citep{ilharco2023editing} is a widely adopted framework for model merging, which builds a multi-task model by linearly combining task vectors. A task vector is defined as $\tau_i=\theta_{i}-\theta_0$, where $\theta_{i}$ is a model fine-tuned on task $i$ from the pre-trained model $\theta_0$. Such linear combinations have been shown to produce effective multi-task models \citep{ilharco2023editing,ortiz-jimenez2023task}. Recent studies focus on refining the task vectors to reduce interference among tasks \citep{yadav2023ties,yu2024language,gargiulo2025task}. We refer to these modified task vectors as \textit{advanced} task vectors. However, the merged model can be sensitive to the linear combination coefficients of these (advanced) task vectors \citep{yang2024adamerging,tam2026realistic}.  Thus, most merging methods use auxiliary data to select these coefficients \citep{chaves2025weight}, via grid search on held-out data \citep{ilharco2023editing,yadav2023ties,yu2024language} or test-time adaptation on unlabeled test data \citep{yang2024adamerging,touayouch2026divmerge}. Despite these advances, a performance gap remains between merged models and individual models \citep{yang2026model}.

To mitigate this gap, we revisit the current model merging pipeline and ask whether the existing framework fully uses the auxiliary data. Optimizing the coefficients of the linear combination is equivalent to restricting weight optimization to the subspace spanned by the (advanced) task vectors. We refer to this subspace as the \textit{task-vector subspace}. From this perspective, we identify coefficient optimization under task arithmetic as a constrained optimization that implicitly regularizes candidate models to the task-vector subspace. While the empirical success of task arithmetic shows that effective multi-task weights exist within the task-vector subspace, its implicit regularization has not been examined. Thus, the research question is: \textit{For model merging, is the implicit regularization induced by task arithmetic actually useful, or does it exclude better multi-task weights?}

To find out the answer, we investigate direct optimization in the weight space beyond the task-vector subspace. We apply the most common optimization method, gradient descent, in standard model merging benchmarks across computer vision and natural language processing.
Surprisingly, we find that optimization in the weight space consistently and significantly improves the performance of various model-merging methods beyond optimizing the coefficients, as shown in \cref{fig:fig1}.
We further downsample the auxiliary dataset and find that the conclusion still holds when data are scarce, with only one sample per class. The results confirm that the optimization flexibility does not lead to overfitting in existing benchmarks.
Moreover, even gradient descent directly from the pretrained model can outperform some existing merging methods. Nevertheless, merged methods still provide stronger initializations for gradient descent. These findings imply that the task arithmetic's implicit regularization may lead to the underuse of auxiliary data and limit merged models' performance. 


We further analyze different strategies from several angles and show that generalization is a key consideration in general cases. First, we confirm that final weights from gradient descent indeed lie outside the task-vector subspace.
Motivated by this finding and  \citet{gan2026neural}, we propose \textit{S-RandOpt} that samples weights outside the subspace. By allowing just one additional degree of freedom, \textit{S-RandOpt} outperforms coefficient optimization, further supporting the idea that the implicit regularization of task arithmetic overly restricts model complexity. It also implies that finding better multi-task weights does not rely on gradient descent. We also investigate the memory and computational costs of different strategies for using auxiliary data.
Finally, we study different strategies for using the auxiliary data under various conditions, revealing the \textbf{bias-variance tradeoff} as a key principle for determining how auxiliary data should be used. Our analysis motivates future merging methods to provide strong initializations that enable more effective use of auxiliary data and, more broadly, calls for rethinking how to build a strong multi-task model via model merging.

\section{Overview of model merging methods} \label{sec: existing works}

The goal of model merging is to build a multi-task model by leveraging models that are separately fine-tuned on different tasks. In this work, we focus on merging models that share the same architecture and are fine-tuned from the same pretrained model on different tasks. While multi-task learning \citep{ruder2017overview} and multi-task fine-tuning \citep{chung2024scaling} also aim to build multi-task models, model merging focuses on the specific scenario where task-specific models are available. This work centers around the model merging scenario, but applies the multi-task learning algorithm.
 
Most model merging methods are based on \textit{task vectors} \citep{ilharco2023editing}, defined as $\tau_i=\theta_{i,ft}-\theta_0$, where $\theta_{i,ft}$ is the model weight fine-tuned on the $i$-th task and $\theta_0$ is the pretrained model they 
share. Given $T$ tasks, the merged model of task-wise merging is then formed as
\begin{equation}
\label{eq:merging}
\theta_{\text{merged}} = \theta_0 + \sum_{i=1}^{T} \lambda_i \, \tilde{\tau}_i. 
\end{equation}

Prior works build better merged models by transforming task vectors into variants $\tilde{\tau}_i$, which we call \textit{advanced task vectors}. One branch of existing methods constructs advanced task vectors by leveraging information from the original task vectors, including sparsification \citep{yadav2023ties,yu2024language} and low-rank decomposition \citep{gargiulo2025task,marczak2025no}.
Another branch of methods requires data, including calculating gradients to estimate curvature information \citep{matena2022merging,daheim2024model}, or estimating activations to reduce the conflict between tasks \citep{yang2024representation}. We further discuss global-wise and layer-wise merging in Appendix \ref{sec:detailed_related_works}.

After building the (advanced) task vectors, the next step is to determine the coefficients $\{\lambda_i\}$ of each task vector. As shown in prior work \citep{yang2024adamerging,tam2026realistic}, merging methods' performance is highly sensitive to these coefficients. The simplest baseline for finding good coefficients is grid search \citep{ilharco2023editing}, which uses a held-out dataset to select the best set of coefficients. However, when merging many tasks, grid search in the $T$-dimensional space requires extensive inference, making it impractical. Therefore, most existing works instead perform diagonal search, using a single shared coefficient for all task vectors.  Another line of work develops methods to optimize the coefficients using this held-out dataset, including gradient descent on coefficients \citep{yang2024adamerging,touayouch2026divmerge} and Bayesian optimization \citep{lee2025dynamic}. Notably, optimizing the coefficients is equivalent to optimizing the model within the subspace spanned by (advanced) task vectors. In this work, we ask if this is the best way to leverage this held-out dataset.

\begin{table}[t]
    \centering
    \caption{\textbf{Summary of prior works in terms of compute and data usage.} \textbf{Backprop.} refers to backpropagation. \textbf{Data Source} refers to the source of auxiliary data. Existing works leverage backpropagation and auxiliary data to build merged models.}
    \label{tab:summary of prior works}
    \renewcommand{\arraystretch}{1.15}
    \begin{tabular}{lcccc}
    \toprule
    &
    \multirow[c]{2}{*}{\raisebox{-2pt}{\textbf{Backprop.}}}
    &
    \multirow[c]{2}{*}{\raisebox{-2pt}{\makecell{\textbf{Data}\\\textbf{Source}}}}
    &
    \multicolumn{2}{c}{\textbf{Auxiliary Data}}
    \\[-3pt]
    \cmidrule(lr){4-5}
    & & &
    \textbf{Inputs}&\textbf{Labels}
    \\
    \midrule
    
    TA  \citep{ilharco2023editing}    & $\times$ & Valid & $\checkmark$ & $\checkmark$ \\
    \rowcolor{gray!20} Fisher averaging \citep{matena2022merging} & $\checkmark$ & Train & $\checkmark$ & $\times$ \\
     TIES  \citep{yadav2023ties}    & $\times$ & Valid & $\checkmark$ & $\checkmark$ \\
    \rowcolor{gray!20} Uncertainty-based \citep{daheim2024model} & $\checkmark$ & Train & $\checkmark$ & $\checkmark$ \\
    DARE  \citep{yu2024language}    & $\times$ & Valid & $\checkmark$ & $\checkmark$ \\
    \rowcolor{gray!20} TSV-M  \citep{gargiulo2025task}    & $\times$ & $\times$ & $\times$ & $\times$\\ \midrule
    AdaMerging \citep{yang2024adamerging}   & $\checkmark$ & Test & $\checkmark$ & $\times$ \\
    \rowcolor{gray!20} DivMerge \citep{touayouch2026divmerge}   & $\checkmark$ & Test & $\checkmark$ & $\times$ \\

    \bottomrule
    \end{tabular}
\end{table}

\subsection{Compute and data usage of existing model merging methods}

As the goal of model merging is to build a multi-task model without retraining on the whole training data, the computational cost and data usage are practical considerations. We collectively refer to additional data used for different purposes as \textit{auxiliary data}, including the held-out (valid) dataset and unlabeled test set used for coefficient optimization, as well as the data used to build advanced task vectors. Adapted from \citet{gargiulo2025task}, \cref{tab:summary of prior works} summarizes the existing model merging methods. Prior work has separately used backpropagation on model weights and auxiliary data to optimize merging coefficients, so we investigate using auxiliary data to optimize model weights.



\section{Does optimizing within task-vector subspace suffice?} \label{sec: main results}
In this section, we compare optimization within the task-vector subspace and the broader weight space. Under standard model-merging benchmarks in vision and language, we evaluate the performance of merged models from common model-merging methods under two optimization choices.

\subsection{Experimental setup} \label{sec: exp setup}
\textbf{Data.} Following the original setup \citep{ilharco2023editing}, we split $10\%$ of the training set as the auxiliary data, and leave the rest as the training data for fine-tuning the task-specific model. We evaluate 9 vision tasks: MNIST \citep{lecun1998mnist}, GTSRB \citep{stallkamp2011german}, FER2013 \citep{goodfellow2013challenges}, Stanford Cars \citep{krause20133d}, DTD \citep{cimpoi2014describing}, Food101 \citep{bossard2014food}, SUN397 \citep{xiao2016sun}, RESISC45 \citep{cheng2017remote}, and EuroSAT \citep{helber2019eurosat}. For language, we evaluate 4 tasks: DDXPlus \citep{tchango2022ddxplus}, Banking77 \citep{loukas2023making}, IFEval \citep{zhou2023instruction}, and Usefulness Judge \citep{chen2026expected}. 


\textbf{Models.} We evaluate 3 CLIP-ViT variants \citep{radford2021learning} for vision, Qwen3-0.6B, 1.7B and 4B \citep{yang2025qwen3} and Llama-3.2-1B \citep{grattafiori2024llama} for language. Training details of task-specific models are provided in Appendix \ref{sec:finetune}.

\textbf{Model merging methods.} We test 4 common model merging methods: vanilla Task Arithmetic, which we refer to as TA \citep{ilharco2023editing}, TIES \citep{yadav2023ties}, DARE \citep{yu2024language} and TSV-M \citep{gargiulo2025task}. \cref{sec: existing works} provides a detailed introduction to these methods.

\subsubsection{Two ways for optimizing merged models}

\textbf{Coefficient search and subspace best.} Existing methods search for good coefficients $\{\lambda_i\}$ in the final step. In this work, we denote \textit{Naive} as setting all coefficients $\lambda_i$ to $\frac{1}{T}$ or $1$, and \textit{Naive baseline} as the best between the two. \textit{Coefficient search} refers specifically to diagonal coefficient search in this work. To estimate the empirical best performance achievable within the task-vector subspace using auxiliary data, we report the best result among three coefficient optimization methods (denoted as \textit{subspace best}): gradient descent on coefficients, fine-grained grid search, and Bayesian optimization. See Appendix \cref{sec:subspace best} for details. We simply refer to the task-vector subspace as the \textit{subspace} in this section, where optimizing the coefficients is equivalent to optimizing the model weights within this subspace. Therefore, we call gradient descent on the coefficients \textit{Subspace GD}.

\textbf{Optimizing model weights.} We minimize the multi-task loss with gradient descent (denoted as \textit{GD}), initialized from the \textit{Naive baseline}. Specifically, we fine-tune the model using auxiliary data from each task. For the main results, we use AdamW \citep{loshchilov2018decoupled}, and discuss hyperparameters in Appendix \ref{sec:training_details}. We also evaluate SGD in Appendix \ref{sec:sgd}. We perform full-weight fine-tuning for vision models and LoRA fine-tuning \citep{hu2022lora} for LLMs.  

\begin{figure}[t]
    \centering
    \includegraphics[width=1\linewidth]{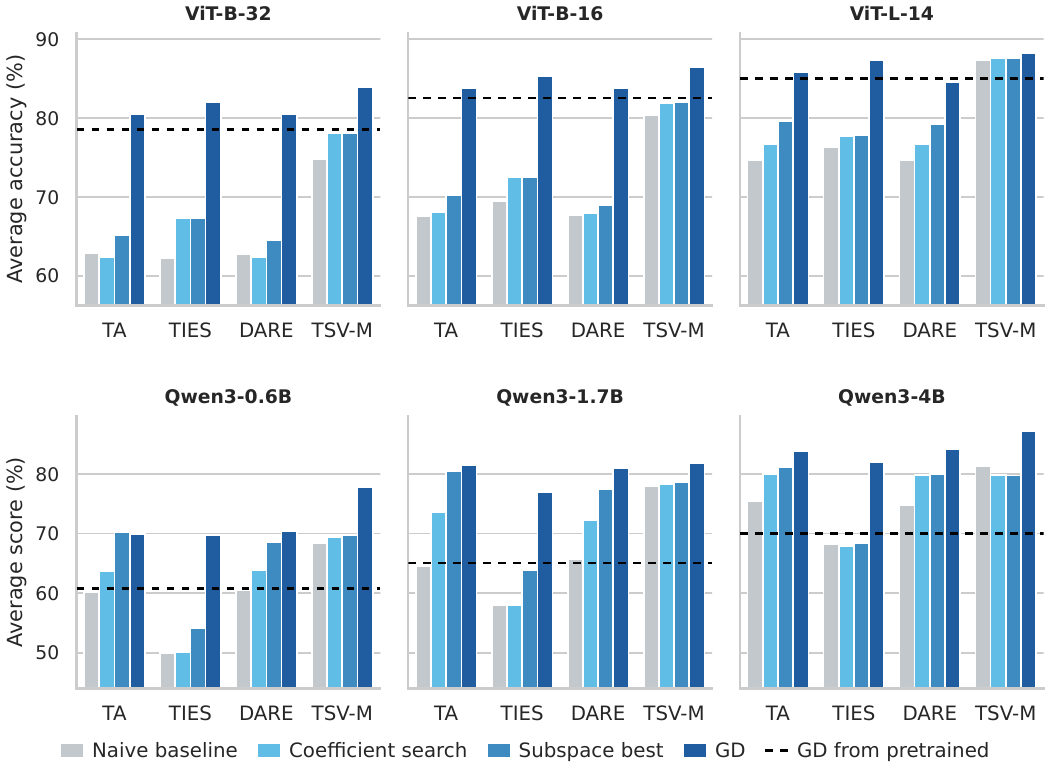}
    \caption{\textbf{Average test accuracy under different uses of auxiliary data.} Across vision tasks (top) and language tasks (bottom) with different models, Gradient Descent (GD) from the merged models consistently outperforms both coefficient search and the empirical best performance of its search space (Subspace best) for every merging method (TA, TIES, DARE, and TSV-M).}
    \label{fig: main}
\end{figure}

\begin{figure}[t]
    \centering
    \includegraphics[width=1\linewidth]{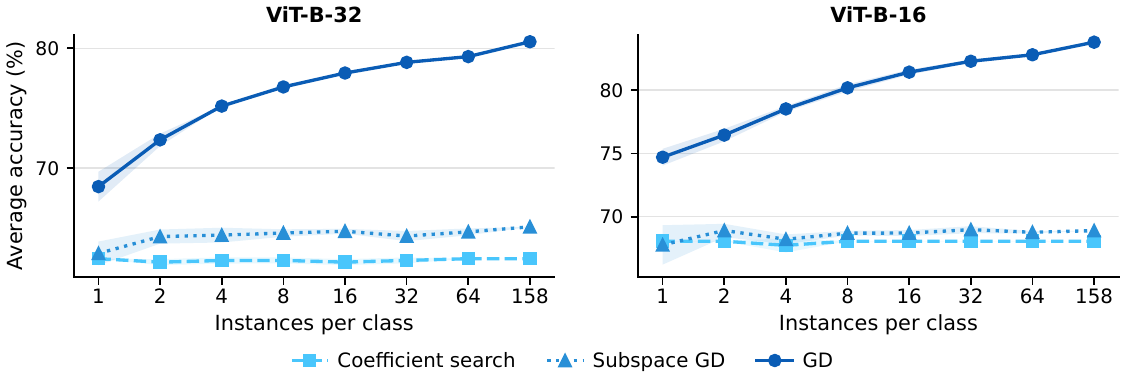}
    \caption{\textbf{Average test accuracy under different sizes and uses of auxiliary data for TA.} We downsample the auxiliary data and compare different uses of auxiliary data for TA. We run three random seeds for downsampling, and report error bars as the shaded regions. Gradient Descent (GD) consistently outperforms the other two approaches that optimize weights within the subspace.}
    \label{fig: data scaling ta}
\end{figure}

\subsection{Results and discussions}

\textbf{All methods benefit from optimizing coefficients.} \cref{fig: main} shows that optimizing the coefficients of all merged methods improves their performance. Meanwhile, optimizing the coefficient beyond the diagonal search can further boost performance, which is consistent with existing work \citep{yang2024adamerging,touayouch2026divmerge}. Notably, coefficient optimization improves every method, including TSV-M, which is relatively insensitive to coefficient choices \citep{gargiulo2025task}.

\textbf{Optimizing weights is better than optimizing coefficients.}  \cref{fig: main} shows that gradient descent (GD) on weights consistently outperforms the subspace's best performance and improves all merged models. While TSV-M does not require auxiliary data for coefficient tuning \citep{gargiulo2025task}, GD still improves its performance when such data are available. We further discuss the correlation between each merging method's naive performance and its performance after GD in \cref{sec: init}.

\textbf{GD from the pretrained model is surprisingly good.} For vision tasks (\cref{fig: main} first row), GD from the pretrained model (GD from pretrained) is even better than TA, TIES, and DARE with optimized coefficients. These results support our claim that the use and importance of auxiliary data have been underestimated. Nonetheless, GD from base is still worse than GD from the merged model, which demonstrates the effectiveness of merging methods.

\textbf{Conclusions hold with limited data.}  We downsample the auxiliary dataset and report vision results for TA in \cref{fig: data scaling ta}; TIES results are in Appendix \ref{sec:more data scaling}.
Weight-space gradient descent still outperforms optimization within the task-vector subspace. This implies that in standard model merging benchmarks, optimizing flexibility does not lead to worse performance even when the data are scarce. Another observation is that weight-space gradient descent better leverages auxiliary data: its performance continues to improve as more data become available, whereas coefficient optimization quickly saturates. We further challenge the limits of gradient descent in \cref{sec: extreme case}.

\section{Leaving the task-vector subspace improves performance} \label{sec: leaving subspace}

\cref{sec: main results} shows that directly optimizing the model weights with gradient descent consistently outperforms coefficient optimization within the task-vector subspace. In this section, we confirm the source of this improvement. In \cref{sec:gd_analysis}, we first examine the geometric relationship between the solutions found by gradient descent and the task-vector subspace. In \cref{sec:srandopt}, we propose another method showing that this improvement does not depend on gradient descent. 


\subsection{Analysis of gradient descent for model merging} \label{sec:gd_analysis}

\begin{figure}[ht]
    \centering
    \begin{minipage}[t]{0.56\textwidth}
        \centering
        \vspace{0pt}
        \includegraphics[width=\linewidth]{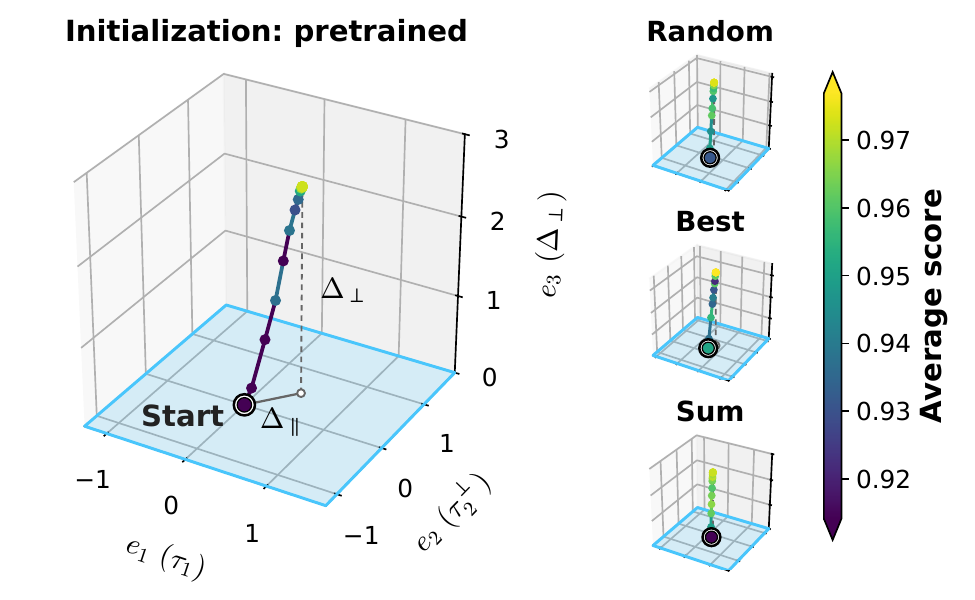}
        \captionof{figure}{\textbf{Gradient descent trajectories in the 2-task setting (ViT-B-32, EuroSAT and GTSRB).} Each panel shows one run initialized on the task-vector subspace (blue plane). The axes $e_1$, $e_2$, and $e_3$ are the directions of $\tau_1$, $\tau_2^{\perp}$, and each run's update's orthogonal component to the subspace $\Delta_\perp$. In all runs, gradient descent leaves the subspace as the score (accuracy) increases.}
        \label{fig:gd_trajectory}
    \end{minipage}
    \hfill
    \begin{minipage}[t]{0.43\textwidth}
        \centering
        \vspace{0pt}
        \captionof{table}{\textbf{Decomposition of the gradient descent update (ViT-B-32, 9 tasks).} We decompose the fine-tuning update $\Delta = \theta_{\text{GD}} - \theta_{\text{merge}}$ into its projection onto the task-vector subspace, $\Delta_\parallel$, and the orthogonal component, $\Delta_\perp= \Delta - \Delta_\parallel$. Adding only $\Delta_\parallel$ to the merged model yields no accuracy gain (Acc. Gain), whereas the full update improves accuracy by 9 to 20 points. The empirical results indicate that the performance improvement requires leaving the task-vector subspace.}
        \label{tab:gradient_analysis_vit32}
        \smallskip
        \resizebox{\linewidth}{!}{
        \begin{tabular}{lrrr}
        \toprule
        \multirow[c]{2}{*}{\raisebox{-2pt}{\textbf{Method}}}
        &
        \multirow[c]{2}{*}{\raisebox{-2pt}{$\lVert\Delta_{\perp}\rVert/\lVert\Delta_{\parallel}\rVert$}}
        &
        \multicolumn{2}{c}{\textbf{Acc. Gain (\%)}}
        \\[-2pt]
        \cmidrule(lr){3-4}
        & &
        $+\Delta_\parallel$
        &
        $+\Delta_\parallel + \Delta_\perp$
        \\
        \midrule
        TA     & 11.14    & -0.04 & 17.66 \\
        TIES   & 21.60    &  0.07 & 19.81 \\
        DARE   & 14.50    & -0.01 & 17.78 \\
        TSV-M  & 78922.16 &  0.00 & 9.09 \\
        \bottomrule
        \end{tabular}
        }
    \end{minipage}

\end{figure}

To check whether the weights obtained by gradient descent lie outside the task-vector subspace, we first visualize gradient descent in a 2-task setting. 
We run gradient descent from several merged models $\theta_{\text{merge}}$ on this plane, and denote the resulting update by $\Delta = \theta_{\text{GD}} - \theta_{\text{merge}}$. For each run, we project the trajectory onto three axes centered at $\theta_0$: two axes spanning the subspace, $\tau_1$ and $\tau_2^{\perp}$ (the component of $\tau_2$ orthogonal to $\tau_1$), and a third axis $\Delta_\perp$, the component of $\Delta$ orthogonal to the subspace. As shown in \cref{fig:gd_trajectory}, gradient descent consistently leaves the subspace regardless of the initialization. 

We next quantify this behavior in the 9-task setting used in \cref{sec: main results}. For each merging method, we run gradient descent from its merged model and decompose the update into $\Delta_\perp$ and the in-subspace component $\Delta_\parallel$. 
\cref{tab:gradient_analysis_vit32} shows that $\Delta_\perp$ dominates the update. 
Moreover, the in-subspace component alone does not improve performance: adding only $\Delta_\parallel$ to the merged model barely changes the accuracy, 
whereas the full update improves it by up to 20 points. Together with \cref{fig:gd_trajectory}, these results show that the performance gain of gradient descent requires leaving the task-vector subspace.

\subsection{S-RandOpt: a simple method to boost model merging} \label{sec:srandopt}

Gradient descent trajectory analysis shows that better multi-task weights are outside the task-vector subspace. Next, we ask: \textit{Is it easy to find these weights?}  \citet{gan2026neural} show that task-specific model weights are dense around the pretrained model weights, and random sampling can find effective model weights. Specifically, they propose \textit{RandOpt} that randomly samples weights and selects the top-performing weights using training data. Inspired by these findings, we ask whether sampling outside the task-vector subspace can find multi-task weights that outperform those within the subspace. We propose \textit{Subspace-RandOpt (S-RandOpt)}, which uses the auxiliary data to compute the gradient from the task-vector subspace, extends the task-vector subspace by this direction, and then randomly samples within the expanded region. Finally, it uses auxiliary data for evaluation to select the best sampled weight. Appendix \ref{sec:srandopt_details} discusses the implementation details.      

\begin{table}[ht]
\centering
\small
\caption{\textbf{S-RandOpt results across vision and language models.} For two merging methods (TA and DARE), we report the average performance of top-5 sampled weights from S-RandOpt. It frequently outperforms the best performance on the task-vector subspace.}
\label{tab:srandopt}
\resizebox{0.8\linewidth}{!}{
\begin{tabular}{ll | cc | cc}
\toprule
\multicolumn{2}{l|}{}
& \multicolumn{2}{c|}{\textbf{Vision}}
& \multicolumn{2}{c}{\textbf{Language}} \\
\textbf{Method} & \textbf{Strategy}
& \textbf{ViT-B-32} & \textbf{ViT-B-16}
& \textbf{Qwen3-0.6B} & \textbf{Qwen3-1.7B} \\
\midrule

\multirow{2}{*}{TA}
& Subspace best
& 65.10 & 70.24
& 70.20 & 80.51 \\
& S-RandOpt
& 68.76 $\pm$ 0.32 & 70.73 $\pm$ 0.43
& 70.53 $\pm$ 0.55 & 80.22 $\pm$ 1.10 \\
\midrule

\multirow{2}{*}{DARE}
& Subspace best
& 64.55 & 68.97
& 68.58 & 77.41 \\
& S-RandOpt
& 68.66 $\pm$ 0.32 &	70.70 $\pm$ 0.34
& 69.60 $\pm$ 0.58 &	79.53 $\pm$ 0.52 \\
\bottomrule
\end{tabular}
}
\end{table}
\begin{figure}[h]
    \centering
    \includegraphics[width=\linewidth]{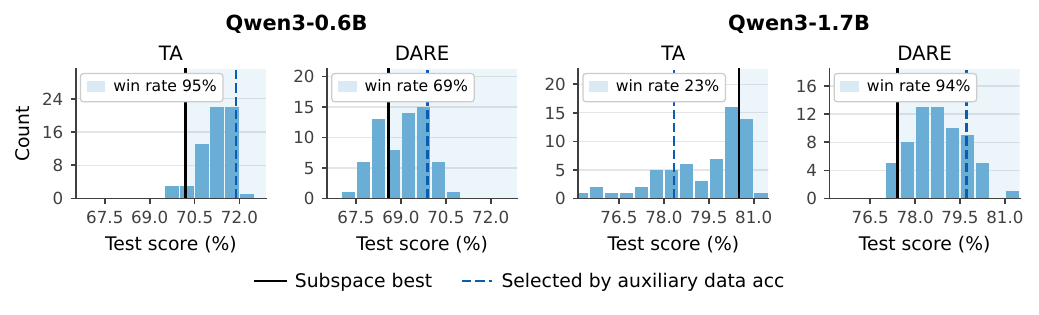}
    \caption{\textbf{S-RandOpt sampled weights' test performance.} We sample $64$ random weights in language models using S-RandOpt, and compare their test performance with the best point on the task-vector subspace (subspace best). A high proportion of sampled weights from outside the subspace outperforms the best weight on the subspace (Subspace best).}
    \label{fig:srandopt_test_distribution}
\end{figure}

As reported in \cref{tab:srandopt}, S-RandOpt frequently outperforms the best weights in the task-vector subspace. Meanwhile, \cref{fig:srandopt_test_distribution} shows that better multi-task weights are dense outside the task-vector subspace. S-RandOpt not only improves the merged models, but also demonstrates that finding better multi-task weights does not depend on gradient descent.

By adding just one additional degree of freedom beyond the task-vector subspace, even random sampling can outperform the best performance in the subspace. These results further confirm that coefficient optimization of task arithmetic might impose unnecessarily strong regularization in model merging benchmarks. As a proof-of-concept, S-RandOpt also motivates future work on developing model-merging algorithms that make better use of auxiliary data by leaving the task-vector subspace.

\section{Analysis for practical use} \label{sec: more analysis}

In this section, we further analyze different strategies for using the auxiliary data, and discuss the practical implications. Specifically, we examine the computational costs of different strategies in \cref{sec:compute}, the role of model merging methods in \cref{sec: init}, the trade-off of the regularization in \cref{sec: extreme case} and \cref{sec: bias-variance tradeoff}, and test-time adaptation of the merged model in \cref{sec:tta}.

\subsection{Computational usage} \label{sec:compute}

\begin{figure}[t]
    \centering
    \begin{subfigure}[t]{0.48\linewidth}
        \centering
        \includegraphics[width=\linewidth]{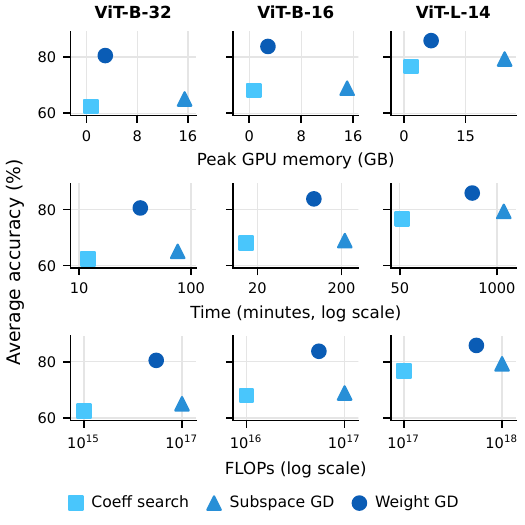}
        \caption{Vision}
        \label{fig: cv ta cost}
    \end{subfigure}
    \hfill
    \begin{subfigure}[t]{0.48\linewidth}
        \centering
        \includegraphics[width=\linewidth]{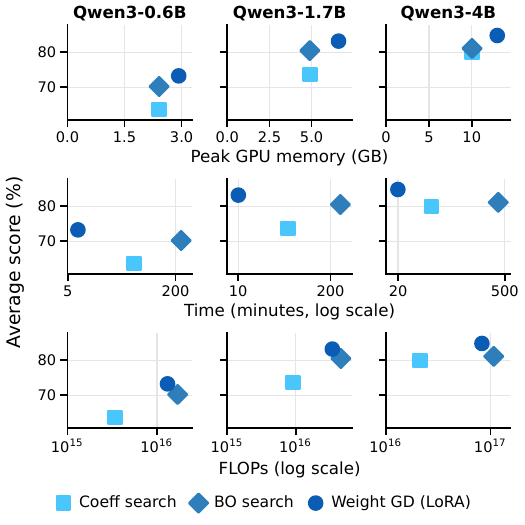}
        \caption{Language}
        \label{fig: llm ta cost}
    \end{subfigure}
    \caption{\textbf{Computational usage of different ways to use the auxiliary data.} Gradient descent on weights (Weight GD) uses slightly more memory than coefficient search. While doing gradient descent requires more compute effort (time/FLOPs), it trades compute for better performance.}
    \label{fig: ta cost}
\end{figure}

As the goal of model merging is to build a multi-task model efficiently,  "training" seems to be an action with high cost. 
To fairly investigate the cost and provide practical recommendations, we analyze different strategies of using auxiliary data from three perspectives: memory, time, and compute. The results are shown in \cref{fig: ta cost}.

\textbf{Memory.} Gradient descent induces larger peak GPU memory usage as it requires backpropagation, while coefficient search and Bayesian optimization only require inference. However, extra memory usage is not huge compared to optimizing coefficients using gradient descent (Subspace GD), where existing work \citep{yang2024adamerging,touayouch2026divmerge} leverages this technique. 

\textbf{Time and Compute.} Gradient descent trades compute for better performance. For vision models, both time and FLOPs follow the same pattern: Subspace GD $>$ Weight GD $>$ Coefficient search. For language models, gradient descent is more time-efficient because coefficient search and Bayesian optimization require repeated inference rollouts with autoregressive decoding.

In summary, gradient descent requires slightly more GPU memory in the standard implementation \footnote{We use standard techniques for GD without substantially compromising speed. See \cref{sec:llm_efficiency_details} for details.}, and trades compute for better performance. We tested SGD in Appendix \ref{sec:sgd} to reduce the GPU memory usage. We recommend that practitioners select the strategy and the compute effect to achieve the Pareto frontier in each scenario.

\subsection{Merging methods provide strong initialization} \label{sec: init}

\begin{wrapfigure}{r}{0.48\textwidth}
    \vspace{-10pt} 
    \centering
     \includegraphics[width=0.8\linewidth]{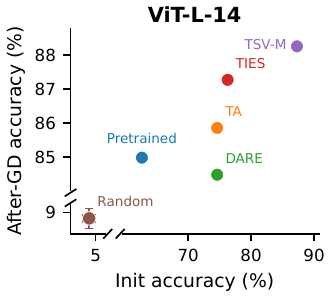}
     \caption{\textbf{Positive correlation between Init accuracy of merged models and their after-GD accuracy.} 
     Existing model merging methods provide strong initializations for gradient descent, and stronger methods have better after-GD performance. 
     }
     \label{fig:init}
     \vspace{-12pt}
\end{wrapfigure}

While \cref{sec: main results} suggests that gradient descent is the better way to use auxiliary data, this finding does not dismiss the importance of model merging methods. \cref{fig:init} shows the results of ViT-L-14, and Appendix \cref{app: init matters} provides more results of other vision and language models. The empirical results show that merging methods with higher initial accuracy also tend to achieve better performance after Gradient Descent (GD). Meanwhile, merging methods provide necessary initialization points for gradient descent, since random initialization yields weak performance. Moreover, this clearly distinguishes this work's contribution from simple multi-task learning or fine-tuning, as starting from merged models performs significantly better than starting from the pretrained model or random initializations.
This result motivates future work on stronger model merging methods to provide stronger initialization for gradient descent and other strategies for using auxiliary data.

\subsection{Extreme cases and underlying concepts} \label{sec: extreme case}

\begin{wrapfigure}{r}{0.48\textwidth}
    \vspace{-8pt}
    \centering
    \includegraphics[width=\linewidth]{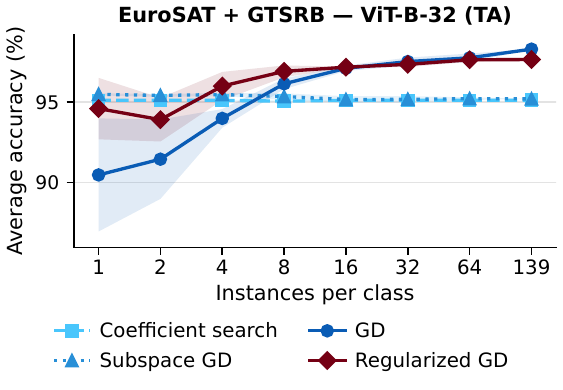}
    \caption{\textbf{Auxiliary data scaling for TA in 2-task setting}. The number of classes (53) is smaller than in the 9-task setting (856). 
    Gradient descent's performance is worse than coefficient search when data is extremely limited, 
    but still outperforms it when more data are provided. Adding regularization for GD mitigates the issue. 
    The results indicate that generalization is a key factor when using auxiliary data.}
    \label{fig: 2task data scaling ta}
    \vspace{-8pt}
\end{wrapfigure}

To understand the big picture, we analyze when gradient descent fails in model merging. \cref{fig: 2task data scaling ta} shows that in the 2-task setting, where the auxiliary data size is smaller than in the 9-task setting in \cref{sec: main results}, GD can perform worse than coefficient search when data are downsampled. As more data are provided, GD eventually outperforms coefficient search. Adding regularization for GD makes low-data scenarios better. However, it slightly decreases performance when data are sufficient. More analysis is in Appendix \ref{sec:2-task-scenario}.

Together with results of coefficient search versus weight-space fine-tuning (\cref{sec: main results}), S-RandOpt (\cref{sec:srandopt}), and full fine-tuning versus LoRA fine-tuning (Appendix \ref{sec:lora}), these results reflect a machine learning principle: \textbf{bias-variance tradeoff} \citep{hastie2009elements}. With less data available, stronger regularization (higher bias) prevents overfitting; with more data available, variance decreases, so greater model flexibility has less risk of overfitting. Coefficient optimization allows limited degrees of freedom and thus imposes stronger regularization, whereas weight-space fine-tuning allows greater flexibility. On standard model merging benchmarks, we have shown that gradient descent does not severely overfit. We recommend practitioners consider the generalization trade-off when deciding how to use auxiliary data.


\subsection{Bias-variance tradeoff} \label{sec: bias-variance tradeoff}

\begin{figure}[h!]
    \centering
\includegraphics[width=\linewidth]{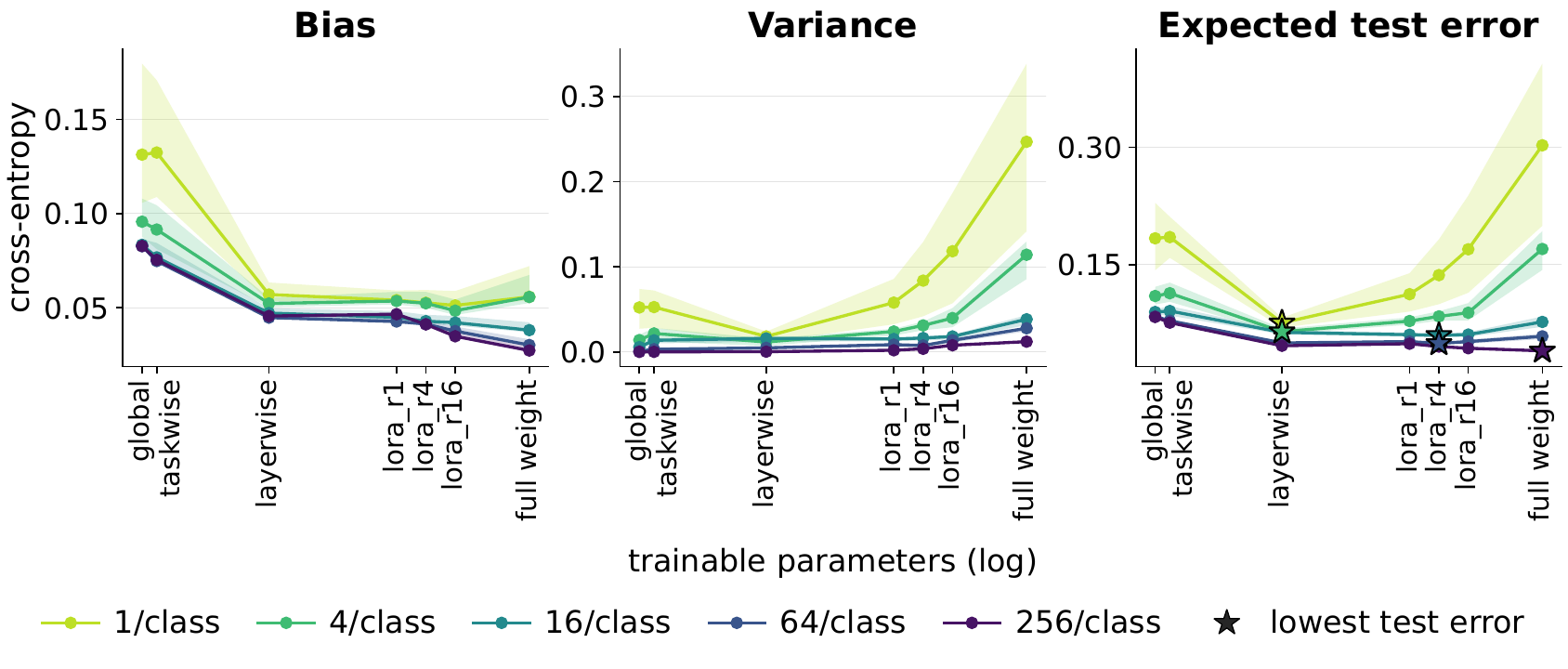}
    \caption{\textbf{Bias-variance decomposition of merging EuroSAT and MNIST on VIT-B-32}. We use the trainable parameters as a proxy for model complexity, and control the instances per class to vary the training data size. The test set is fixed. We test three types of coefficient optimization (global, taskwise, and layerwise merging), LoRA with three ranks ($r=1,4,16$), and full-weight fine-tuning. The results indicate that different training data sizes require different levels of model complexity (regularization strength).}
    \label{fig: bv_tradeoff_eurosat_mnist}
\end{figure}

We now provide a unified analysis of the regularization strengths and test-set performance of all strategies discussed in this paper, based on bias-variance decomposition \citep{hastie2009elements}. We use trainable parameters as a proxy for the model complexity. Coefficient optimization methods, including global-, task-, and layer-wise merging, have only $1$, $T$, and $\text{\# layer}\times T$ parameters, respectively, making them strongly regularized, low-complexity methods. In contrast, LoRA and full-weight fine-tuning stand for weak-regularization methods. Following the previous works \citep{heskes1998bias,yang2020rethinking}, we decompose the expected cross-entropy loss into the bias and variance terms.  We vary the training dataset sizes and sample 20 different training subsets for each size to empirically estimate the expected test error. In this section, we conduct our analysis based on TA. See Appendix \ref{app: bv_tradeoff} for more details and analyses.

The results for merging EuroSAT and MNIST are shown in \cref{fig: bv_tradeoff_eurosat_mnist}. With a small training dataset, more flexible methods (LoRA and full-weight fine-tuning) are more prone to overfitting (higher variance), leading to higher test error. In contrast, less flexible methods (global, task, and layer-wise merging) have higher bias, while the significant lower variance makes them achieve a better bias-variance tradeoff. With a large training dataset, more flexible methods suffer less from overfitting and have lower bias, leading to lower test error. These results show that generalization is a key consideration when choosing the regularization strength for strategies using auxiliary data.

\subsection{Test-time adaptation} \label{sec:tta}

\begin{wraptable}{r}{0.5\linewidth}
    \vspace{-12pt}
    \centering
    \caption{\textbf{Test-time adaptation methods' performance.} All methods are applied on top of TA. Compared with existing methods that use an unlabeled test set to optimize the coefficients of task vectors, using this auxiliary data to perform Gradient Descent (GD) on weights leads to better accuracy across all models.}
    \label{tab:tta}
    \small
    \resizebox{\linewidth}{!}{%
    \begin{tabular}{lccc}
        \toprule
        Method & ViT-B-32 & ViT-B-16 & ViT-L-14 \\
        \midrule
        AdaMerging & 62.21 & 66.80 & 76.63 \\
        DivMerge   & 65.43 & 69.62 & 78.49 \\
        GD  & \textbf{82.43} & \textbf{85.51} & \textbf{88.15} \\
        \bottomrule
    \end{tabular}
    }
\end{wraptable}
One way to address the generalization issue is test-time adaptation \citep{wang2021tent,liang2025comprehensive}. Some previous model merging works \citep{yang2024adamerging, touayouch2026divmerge} have leveraged unlabeled test data to adapt the merged models on test data. However, they only optimize the coefficients. For TA \citep{ilharco2023editing}, \cref{tab:tta} shows that using gradient descent to adapt the unlabeled test data outperforms coefficient optimization. Detailed settings and more results are in Appendix \ref{sec:tta_appendix}, further supporting the bias-variance tradeoff as a key principle for using auxiliary data.

\section{Conclusion}
This work identifies that optimizing weights in the task-vector subspace is a potentially overlooked bottleneck in model merging. Through extensive experiments, we call for a broader look at model merging: reconsider whether the implicit regularization of task arithmetic is useful, and the need to explore beyond the task-vector subspace. Across standard model merging benchmarks, we find that this regularization often limits the merged model's performance. By relaxing the regularization, simple gradient descent or even sampling can improve the merged model's performance. In general, we reveal that generalization is a key factor to consider when determining how strongly the regularization should be applied. That said, we did not identify a one-size-fits-all method for model merging, as the appropriate method depends on the models, auxiliary data, and scenarios. Overall, our findings motivate future work on developing stronger merging methods and model-merging pipelines that jointly consider auxiliary-data usage and optimization flexibility.

\clearpage

\section*{Reproducibility statement}

Experiment details are clearly described in the main paper and Appendix \ref{sec:training_details} to reproduce the empirical results. We have also provided the source code at \url{https://github.com/appier-research/broader_model_merging}. The datasets used are all publicly available.

\section*{Acknowledgments}

We would like to thank Yu-Ang Lee, Ren-Wei Liang, Zhi Rui Tam, and Cheng-Kuang Wu for their feedback on this work. We would also like to thank Zhi Rui Tam for his help with fine-tuning several language models.
This work was supported in part by the National Science and Technology Council, Taiwan, under Grants 114-2628-E-002-021-, 114-2628-E-A49-002, 115-2634-F-002-012-, 115-2223-E-002-005-MY3, and 115-2218-E-002-026-, and the Taiwan Centers of Excellence in Artificial Intelligence. Shao-Hua Sun was supported by the Yushan Fellow Program of the Ministry of Education, Taiwan.

\bibliography{iclr2027_conference}
\bibliographystyle{iclr2027_conference}

\clearpage
\appendix
\section*{Appendix}

\vspace{-1.8cm}
\begingroup
\hypersetup{colorlinks=false, linkcolor=black}
\hypersetup{pdfborder={0 0 0}}
\part{} 
\parttoc 
\endgroup

\section{Training details} \label{sec:training_details}

\subsection{Training individual fine-tuned models} \label{sec:finetune}

\subsubsection{Vision models}
We fine-tune CLIP \citep{radford2021learning} image encoders (ViT-B-32, ViT-B-16, ViT-L-14) separately on each of nine image-classification tasks: DTD \citep{cimpoi2014describing}, EuroSAT \citep{helber2019eurosat}, FER2013 \citep{goodfellow2013challenges}, Food101 \citep{bossard2014food}, GTSRB \citep{stallkamp2011german}, MNIST \citep{lecun1998mnist}, RESISC45 \citep{cheng2017remote},
Stanford Cars \citep{krause20133d}, and SUN397 \citep{xiao2016sun}. Following \citet{ilharco2023editing}, the classification head of each task is the CLIP zero-shot head: we encode every class name with the task's prompt templates using the CLIP text encoder, and the (logit-scaled) normalized text embeddings serve as the weights of a linear layer with zero bias. Both the text encoder and this head stay frozen, so only the image encoder is updated, and every task vector lives in the same parameter space. We minimize the cross-entropy loss with AdamW (learning rate $10^{-5}$, no weight decay, batch size 32) for 10{,}000 steps. The learning rate warms up linearly over the first 10\% of steps and then decays linearly. Images are preprocessed with the standard CLIP processor and no data augmentation.

\subsubsection{Language models}
We fully fine-tune four base models: Qwen3-0.6B-Base, Qwen3-1.7B-Base, Qwen3-4B-Base \citep{yang2025qwen3}, and Llama-3.2-1B-Instruct \citep{grattafiori2024llama}, on each of four tasks: Banking77 \citep{loukas2023making}, DDXPlus \citep{tchango2022ddxplus}, IFEval \citep{zhou2023instruction}, and Usefulness Judge \citep{chen2026expected}. Training uses Axolotl\footnote{\url{https://github.com/axolotl-ai-cloud/axolotl}} with supervised fine-tuning on chat-formatted data. The loss covers only assistant turns, including the end-of-turn token. We use AdamW ($\beta=(0.9,0.999)$, $\epsilon=10^{-8}$, no weight decay), a peak learning rate of $4\times10^{-5}$ with 10\% linear warmup and cosine decay, an effective batch size of 36, 6 epochs, bf16 precision. The maximum sequence length is 4{,}096 (8{,}192 for DDXPlus). We select the checkpoints with best evaluation accuracies on a held-out set.

\subsection{Data details and split}
\label{app:data-splits}
Each task has three disjoint splits. The \emph{train} split is used only to fine-tune the individual models. The \emph{validation} split is the only data merging strategies can access and is referred to as the auxiliary data in this paper. The \emph{test} split is used only for the final evaluations. All random splits use a fixed seed (42).

\paragraph{Vision.}
We randomly hold out 10\% of each dataset's official training split as the validation set, fine-tune on the remaining 90\%, and test on the complete official test split (Food101 has no test split, so we use its official validation split). In the data-scaling experiments, we subsample $k \in \{1,2,4,8,16,32,64\}$ examples per class from this fixed validation set, using three seeds. When a class has fewer than $k$ instances, we use all instances. The number of classes and samples of each split is listed in \cref{tab:vision-splits}.

\begin{table}[h]
\centering
\caption{Vision data splits.}
\label{tab:vision-splits}
\small
\begin{tabular}{lrrrr}
\toprule
Task & \#Classes & \#Train & \#Valid & \#Test \\
\midrule
DTD           & 47  & 3{,}384  & 376     & 1{,}880 \\
EuroSAT       & 10  & 19{,}440 & 2{,}160 & 2{,}700 \\
FER2013       & 7   & 25{,}838 & 2{,}871 & 7{,}178 \\
Food101       & 101 & 68{,}175 & 7{,}575 & 25{,}250 \\
GTSRB         & 43  & 23{,}976 & 2{,}664 & 12{,}630 \\
MNIST         & 10  & 54{,}000 & 6{,}000 & 10{,}000 \\
RESISC45      & 45  & 17{,}010 & 1{,}890 & 6{,}300 \\
Stanford Cars & 196 & 7{,}330  & 814     & 8{,}041 \\
SUN397        & 397 & 17{,}865 & 1{,}985 & 19{,}850 \\
\bottomrule
\end{tabular}
\end{table}

\paragraph{Language.}
\cref{tab:llm-sources} lists the data source for each task, and \cref{tab:llm-splits} lists the number of samples of each split. Except for IFEval, validation and test are a random split of the same held-out pool. Every task uses a validation:test ratio of 1:9, because weight GD and subspace GD train on the validation set and a larger one would favor them over coefficient search. 

\begin{table}[h]
\centering
\caption{Data sources for the language tasks (Hugging Face dataset IDs). IFEval has no
reference answers, so its fine-tuning data and validation set both come from
\texttt{argilla/ifeval-like-data}. We keep only validation responses that pass strict
prompt-level accuracy, and test on the complete original IFEval benchmark.}
\label{tab:llm-sources}
\small
\begin{tabular}{lll}
\toprule
Task & Dataset & Used for \\
\midrule
\multirow{2}{*}{Banking77}
    & \texttt{theblackcat102/bank77\_m}                    & Train \\
    & \texttt{appier-ai-research/bank-77} (\texttt{in\_domain}) & Valid, Test \\
\midrule
\multirow{2}{*}{DDXPlus}
    & \texttt{appier-ai-research/StreamBench} (\texttt{validate}) & Train \\
    & \texttt{appier-ai-research/StreamBench} (\texttt{test})     & Valid, Test \\
\midrule
\multirow{3}{*}{IFEval}
    & \texttt{argilla/ifeval-like-data} (\texttt{filtered})   & Train, Valid \\
    & \texttt{google/IFEval}              & Test \\
\midrule
\multirow{2}{*}{Usefulness Judge}
    & \texttt{miulab/usefulness-judge} (\texttt{train})    & Train \\
    & \texttt{miulab/usefulness-judge} (\texttt{test})     & Valid, Test \\
\bottomrule
\end{tabular}
\end{table}

\begin{table}[h]
\centering
\caption{Language data splits.}
\label{tab:llm-splits}
\small
\begin{tabular}{lrrr}
\toprule
Task & \#Train & \#Valid & \#Test \\
\midrule
Banking77        & 15{,}004 & 100 & 900 \\
DDXPlus          & 1{,}372  & 176 & 1{,}588 \\
IFEval           & 14{,}448 & 60  & 541 \\
Usefulness Judge & 30{,}732 & 25  & 225 \\
\bottomrule
\end{tabular}
\end{table}

\newpage
\subsection{Subspace GD learning rate sweep and initialization point}
To find the task vector subspace's best performance, we tested different learning rates for subspace GD. \cref{tab:subspace_gd_lr} shows that accuracy is stable across learning rates, and we chose $10^{-2}$ to balance accuracy and speed. We report the subspace GD from coefficient best (Coeff. Best) since it generally has higher performance.
\begin{table}[h]
\centering
\small
\caption{\textbf{Subspace GD with different learning rates on Task Arithmetic and TIES for ViT-B-32.} The performance is stable for learning rate from $10^{-4}$ to $10^{-1}$.}
\label{tab:subspace_gd_lr}
\begin{tabular}{l | cc | cc}
\toprule
\multicolumn{1}{c|}{}
& \multicolumn{2}{c|}{\textbf{Task Arithmetic}}
& \multicolumn{2}{c}{\textbf{TIES}} \\
\textbf{LR/Initialization point}
& \textbf{Avg} & \textbf{Coeff. Best}
& \textbf{Merged} & \textbf{Coeff. Best} \\
\midrule

$10^{-4}$
& \textbf{65.22} & \textbf{65.26}
& 67.05 & 66.75 \\

$10^{-3}$
& 65.00 & 65.07
& 66.73 & 66.71 \\

$10^{-2}$ (def.)
& 65.15 & 64.99
& 66.71 & 66.81 \\

$10^{-1}$
& 65.00 & 64.75
& 66.83 & \textbf{67.03} \\

$1$
& 63.08 & 61.78
& \textbf{67.07} & 66.78 \\

\bottomrule
\end{tabular}
\end{table}

\subsection{Weight GD learning rate sweep and initialization point} \label{sec: gd_lr_init_sweep}
To test the robustness of gradient descent on model weights, we tested different learning rates for GD. Across several settings in \cref{tab:gd_lr} and \cref{tab:budget_gd_lr}, setting learning rate to $10^{-5}$ mostly has best performance. Meanwhile, GD from better initialization (best coefficient found by coefficient search, Coeff. Best) has better performance. In the main paper (\cref{sec: main results}), we report the GD performance from the Naive baseline, so it does not depend on the coefficient search. 

\begin{table}[ht]
\centering
\small
\caption{\textbf{Effect of learning rate across architectures and merging methods for gradient descent on weights.} Gradient descent's initialization points affect the performance.}
\label{tab:gd_lr}
\resizebox{\linewidth}{!}{
\begin{tabular}{l cccc cccc}
\toprule
& \multicolumn{4}{c}{\textbf{ViT-B-32}}
& \multicolumn{4}{c}{\textbf{ViT-B-16}} \\
\cmidrule(lr){2-5}
\cmidrule(lr){6-9}

& \multicolumn{2}{c}{\textbf{Task Arithmetic}}
& \multicolumn{2}{c}{\textbf{TIES}}
& \multicolumn{2}{c}{\textbf{Task Arithmetic}}
& \multicolumn{2}{c}{\textbf{TIES}} \\
\cmidrule(lr){2-3}
\cmidrule(lr){4-5}
\cmidrule(lr){6-7}
\cmidrule(lr){8-9}

\textbf{LR}
& \textbf{Avg} & \textbf{Coeff. Best}
& \textbf{Avg} & \textbf{Coeff. Best}
& \textbf{Avg} & \textbf{Coeff. Best}
& \textbf{Avg} & \textbf{Coeff. Best} \\
\midrule

$10^{-6}$
& 79.52 & 79.41
& 77.82 & 81.25
& 82.58 & 83.07
& 81.18 & \textbf{84.94} \\

$10^{-5}$ (def.)
& \textbf{80.61} & \textbf{80.37}
& \textbf{79.49} & \textbf{82.08}
& \textbf{83.42} & \textbf{84.14}
& \textbf{82.81} & 84.71 \\

$10^{-4}$
& 71.98 & 71.41
& 71.54 & 72.35
& 69.24 & 69.29
& 69.03 & 68.96 \\

\bottomrule
\end{tabular}
}
\end{table}
\begin{table}[ht]
\centering
\small
\caption{\textbf{Effect of learning rate across different optimization budgets for gradient descent on weights.} Same conclusion with the \cref{tab:gd_lr}.}
\label{tab:budget_gd_lr}
\resizebox{0.7\linewidth}{!}{
\begin{tabular}{ll | cc | cc}
\toprule
\multicolumn{2}{l|}{}
& \multicolumn{2}{c|}{\textbf{ViT-B-32}}
& \multicolumn{2}{c}{\textbf{ViT-B-16}} \\
\textbf{Budget} & \textbf{LR}
& \textbf{Avg} & \textbf{Coeff. Best}
& \textbf{Avg} & \textbf{Coeff. Best} \\
\midrule

\multirow{3}{*}{1 / class}
& $10^{-6}$
& \textbf{69.12} & \textbf{68.32}
& 73.65 & 75.35 \\
& $10^{-5}$ (def.)
& 67.02 & 66.17
& \textbf{73.83} & \textbf{75.91} \\
& $10^{-4}$
& 49.41 & 47.49
& 52.64 & 50.32 \\

\midrule

\multirow{3}{*}{4 / class}
& $10^{-6}$
& 74.03 & 75.74
& 77.44 & 79.41 \\
& $10^{-5}$ (def.)
& \textbf{75.03} & \textbf{76.89}
& \textbf{78.24} & \textbf{80.37} \\
& $10^{-4}$
& 63.45 & 64.37
& 59.21 & 61.17 \\

\midrule

\multirow{3}{*}{16 / class}
& $10^{-6}$
& 76.92 & 76.66
& 80.22 & 81.85 \\
& $10^{-5}$ (def.)
& \textbf{77.98} & \textbf{77.66}
& \textbf{81.35} & \textbf{82.54} \\
& $10^{-4}$
& 68.79 & 68.91
& 68.12 & 68.77 \\

\bottomrule
\end{tabular}
}
\end{table}

\newpage
\subsection{Different optimizers} \label{sec:sgd}
In this section, we analyze whether gradient descent for model merging depends on specific optimizers. As shown in \cref{tab:sgd_vs_adamw}, optimizing using SGD does not decrease the performance much. Since AdamW is more efficient than SGD, we report AdamW's results in the main paper. Nonetheless, practitioners could choose SGD if the GPU memory size is limited. 
\begin{table}[h]
\centering
\small
\caption{\textbf{Comparison of weight-space optimizers across different data-size merging methods for ViT-B-32.} Coefficient search and Subspace GD are provided for reference. The empirical results show that Weight GD does not depend on specific optimizers. Meanwhile, they mostly outperform coefficient search and subspace GD, which optimize within the task vector subspace.}
\label{tab:sgd_vs_adamw}
\resizebox{\linewidth}{!}{
\begin{tabular}{l cccccccc c}
\toprule
& \multicolumn{8}{c}{Task Arithmetic}
& \multicolumn{1}{c}{TIES} \\
\cmidrule(lr){2-9}
\cmidrule(lr){10-10}

Method / Data size
& 1 / class
& 2 / class
& 4 / class
& 8 / class
& 16 / class
& 32 / class
& 64 / class
& Full
& Full \\
\midrule

Coefficient search
& 62.41
& 61.97
& 61.97
& 62.41
& 62.41
& 62.41
& 62.41
& 62.41
& 67.35 \\

Subspace GD
& 63.74
& 64.27
& 65.06
& 64.88
& 64.94
& 64.72
& 64.71
& 65.08
& 66.73 \\

Weight GD (AdamW)
& 67.23
& 72.05
& 75.17
& 76.93
& 77.97
& 78.85
& 79.43
& 80.53
& 82.07 \\

Weight GD (SGD)
& 62.17
& 68.86
& 72.53
& 75.23
& 76.68
& 77.66
& 78.62
& 79.68
& 81.79 \\

\midrule

SGD $-$ AdamW
& $-5.06$
& $-3.19$
& $-2.64$
& $-1.70$
& $-1.29$
& $-1.19$
& $-0.81$
& $-0.85$
& $-0.28$ \\

\bottomrule
\end{tabular}
}
\end{table}

\clearpage

\subsection{LLM efficiency implementation details}
\label{sec:llm_efficiency_details}

A naive implementation of full-weight gradient descent (GD) can require prohibitive GPU memory for modern large language models. Standard techniques substantially reduce this cost with little loss in speed. We use two of them, gradient checkpointing \citep{chen2016training} and a fused linear cross-entropy kernel \citep{hsu2025ligerkernel}, and apply them identically to every gradient-based strategy (full-weight, LoRA, and subspace GD). They therefore reduce the memory of all GD methods equally and leave their relative ordering unchanged.

\textbf{Gradient checkpointing}\footnote{Hugging Face implementation: \url{https://huggingface.co/docs/transformers/grad_checkpointing}} stores only the input hidden state of each decoder layer during the forward pass and recomputes the remaining activations as needed during the backward pass. For a batch of $B$ sequences of length $L$ and a model with hidden size $H$, this reduces activation memory from all per-layer intermediates to a single $B \times L \times H$ tensor per layer, at the cost of one extra forward pass per step. We include this cost explicitly in the reported FLOPs (backward $= 2\times$ forward $+\ 1\times$ recomputation).

\textbf{Fused linear cross-entropy}\footnote{Official implementation: \url{https://github.com/linkedin/Liger-Kernel}. We use only \texttt{LigerFusedLinearCrossEntropyLoss}; the rest of the forward pass is identical to the Hugging Face implementation.} avoids materializing the $B \times L \times |V|$ logits tensor, where $|V|$ is the vocabulary size. The default next-token loss keeps this tensor in fp32, along with its log-softmax and gradient, until the backward pass. For Qwen3 ($|V| = 151{,}936$), it occupies about 3.7\,GiB per 2048-token sequence, regardless of model size. The fused kernel instead computes the output projection, log-softmax, and gradient in chunks directly from the final hidden states and the \texttt{lm\_head} weight. It produces the same loss and gradients (within $10^{-4}$ relative error in fp32) with no extra FLOPs. Decoding already enjoys an analogous saving by default, since generation computes logits only at the last position. Applying the kernel during training therefore restores parity between search-based and gradient-based strategies rather than favoring the latter.

Apart from these two techniques, we use no optimization that trades speed for memory. Every strategy runs with the same per-step batch size and keeps its full working set on a single GPU, so the reported peak memory reflects what each method requires rather than the hardware it ran on.

\subsection{Hyperparameters of strategies}

\cref{tab:hyper_vis} and \cref{tab:hyper_language} provide the hyperparameters for the \cref{sec: main results}. 

\begin{table}[ht]
\centering
\caption{Hyperparameters used for vision experiment.}
\label{tab:hyper_vis}
\resizebox{0.55\linewidth}{!}{
\begin{tabular}{lccc}
\toprule
Hyperparameter
& ViT-B-32
& ViT-B-16
& ViT-L-14 \\
\midrule

Batch size
& 32
& 8
& 4  \\

\midrule
Coefficient search $\lambda_{\min}$
& 0.0
& 0.0
& 0.0 \\

Coefficient search $\lambda_{\max}$
& 1.0
& 1.0
& 1.0 \\

Coefficient search steps
& 11
& 11
& 11 \\

\midrule
Weight GD optimizer
& AdamW
& AdamW
& AdamW \\

Weight GD learning rate
& $10^{-5}$
& $10^{-5}$
& $10^{-5}$ \\

Weight GD epochs
& 10
& 10
& 10 \\

Weight GD warmup ratio
& 0.1
& 0.1
& 0.1 \\

Weight GD scheduler
& cosine
& cosine
& cosine \\

Weight GD patience
& 5
& 5
& 5 \\

\midrule
Subspace GD optimizer
& AdamW
& AdamW
& AdamW \\

Subspace GD learning rate
& $10^{-2}$
& $10^{-2}$
& $10^{-2}$ \\

Subspace GD epochs
& 20
& 20
& 20 \\

Subspace GD warmup ratio
& 0.1
& 0.1
& 0.1 \\

Subspace GD scheduler
& cosine
& cosine
& cosine \\

Subspace GD patience
& 5
& 5
& 5 \\

\midrule
TIES density
& 0.2
& 0.2
& 0.2 \\

DARE drop rate
& 0.5
& 0.5
& 0.5 \\

\bottomrule
\end{tabular}
}
\end{table}

\begin{table}[ht]
\centering
\caption{Hyperparameters used for language models.}
\label{tab:hyper_language}
\begin{tabular}{lccc}
\toprule
Hyperparameter
& Qwen3-0.6B
& Qwen3-1.7B
& Qwen3-4B \\
\midrule

Coefficient search $\lambda_{\min}$
& 0.1
& 0.1
& 0.1 \\

Coefficient search $\lambda_{\max}$
& 1.0
& 1.0
& 1.0 \\

Coefficient search steps
& 10
& 10
& 10 \\

\midrule
Weight GD optimizer
& AdamW
& AdamW
& AdamW \\

Weight GD learning rate
& $3 \times 10^{-5}$
& $3 \times 10^{-5}$
& $3 \times 10^{-5}$ \\

Weight GD epochs
& 5
& 5
& 5 \\

Weight GD warmup ratio
& 0.1
& 0.1
& 0.1 \\

Weight GD scheduler
& cosine
& cosine
& cosine \\

Weight GD patience
& none
& none
& none \\

Weight GD micro-batch size
& 1
& 1
& 1 \\

Weight GD grad accum steps
& 4
& 4
& 4 \\

Weight GD effective batch size
& 4
& 4
& 4 \\

Weight GD max sequence length
& 2048
& 2048
& 2048 \\

Weight GD LoRA learning rate
& $3 \times 10^{-4}$
& $3 \times 10^{-4}$
& $3 \times 10^{-4}$ \\

LoRA rank
& 16
& 16
& 16 \\

LoRA alpha
& 32
& 32
& 32 \\

LoRA dropout
& 0.05
& 0.05
& 0.05 \\

\midrule
Subspace GD optimizer
& AdamW
& AdamW
& AdamW \\

Subspace GD learning rate
& $10^{-2}$
& $10^{-2}$
& $10^{-2}$ \\

Subspace GD epochs
& 5
& 5
& 5 \\

Subspace GD warmup ratio
& 0.1
& 0.1
& 0.1 \\

Subspace GD scheduler
& cosine
& cosine
& cosine \\

Subspace GD micro-batch size
& 1
& 1
& 1 \\

Subspace GD grad accum steps
& 4
& 4
& 4 \\

Subspace GD effective batch size
& 4
& 4
& 4 \\

\midrule
Generation temperature
& 0.01
& 0.01
& 0.01 \\

Generation top-$p$
& 0.95
& 0.95
& 0.95 \\

Generation thinking
& off
& off
& off \\

Max new tokens
& \multicolumn{3}{c}{
    \shortstack{
        256 (IFEval), 
        16 (other datasets)
    }
} \\

\midrule
TIES density
& 0.2
& 0.2
& 0.2 \\

DARE drop rate
& 0.5
& 0.5
& 0.5 \\

\bottomrule
\end{tabular}
\end{table}

\clearpage
\section{More related works} \label{sec:detailed_related_works}

In this section, we discuss more details of the related work. 

\subsection{Variants of task-wise merging}

\cref{eq:merging} mainly discusses the \textbf{task-wise merging}, where each task has its own coefficient.  We now discuss other variants.

\textbf{Global-wise merging} produces a single advanced task vector $\tilde{\tau} = f(\{\tau_i\}_{i=1}^T)$. A single coefficient $\lambda \in \mathbb{R}$ is shared across all tasks:
    \begin{equation} \label{eq:global-wise merging}
        \theta_{\text{merged}} = \theta_0 + \lambda \cdot \tilde{\tau}.
    \end{equation}
TIES \citep{yadav2023ties} and TSV-M \citep{gargiulo2025task} belong to this class. Their task vector subspace is a one-dimensional subspace, and optimizing this single coefficient also implicitly regularizes the candidate models into this one-dimensional task-vector subspace. One can rewrite \cref{eq:global-wise merging} into \cref{eq:merging} (task-wise merging). 

\textbf{Layer-wise merging} \citep{yang2024adamerging,sun2025towards,xu2025scalable} further merged parameters at each layer $l$ independently:
\begin{equation}
    \theta_{\text{merge}}^{(l)} = \theta_0^{(l)} + \sum_{i=1}^T \lambda_{i,l} \cdot \tilde{\tau}_{i,l}, \quad \forall l = 1, \ldots, L,
\end{equation}

where $L$ is the number of layers. The final merged model is

\begin{equation}
    \theta_{\text{merge}} = \theta_0 + \sum_{i=1}^T \sum^L_{l=1} \lambda_{i,l} \cdot \tilde{\tau}_{i,l}, 
\end{equation}

where $\tilde{\tau}_{i,l} \in\mathbb{R}^{d}$ denotes the layer-$l$
component of task vector $i$, embedded in the full weight space with $d$
parameters, and can be viewed as a type of advanced task vector. Layer-wise merging uses $L*T$ coefficients, compared with at most $T$ coefficients for task-wise merging. Therefore, it has more trainable parameters (coefficients) than task-wise merging. However, it still allows fewer trainable parameters (coefficients) than full-weight fine-tuning. Essentially, layer-wise merging creates a larger task-vector subspace that regularizes candidate models into this subspace.

Notably, \citet{xu2025scalable} compared their layer-wise merging method with full-weight fine-tuning with auxiliary data. However, they did not specify the initialization of the fine-tuning with gradient descent, and our studies in \cref{sec: main results}, \cref{sec: init}, \cref{sec: gd_lr_init_sweep} and \cref{sec:lora} have demonstrated the importance of the initialization point for gradient descent. However, results in \cref{sec:2-task-scenario} and Appendix \ref{sec:tta_appendix} shows that full-weight fine-tuning can underperform task-wise or layer-wise merging due to overfitting when data are scarce. How strong the regularization should be depends on the auxiliary dataset.

\subsection{Detailed discussion regarding compute and data usage}

Besides the related work discussed in \cref{sec:compute}, we also identify other existing works \citep{tang2024merging,lee2026adarank} doing backpropagation through the model weights. This suggests that, when computational resources allow, practitioners may trade additional computation for better performance. 

We also clarify that model merging methods classified as data-free in prior work \citep{cheng2025whoever,chaves2025weight,yang2026model} are typically insensitive to coefficient choices \citep{gargiulo2025task} or have no tunable coefficients or hyperparameters \citep{cheng2025whoever}. \cref{sec: main results} shows that for methods that have coefficients, carefully tuning them leads to better performance. Even for merging methods without tunable coefficients, auxiliary data can still be used to further improve performance through gradient descent. As shown in \cref{sec: init}, merged models provide strong initialization points for gradient descent. Thus, when auxiliary data are available, further optimization may help close the performance gap between a merged model and individual task-specific models.

\clearpage
\section{Main results details}

\subsection{Finding Subspace Best Performance} \label{sec:subspace best}

We tried three methods to find the best performance on the task-vector subspace by using auxiliary data: gradient descent on coefficients (Subspace GD), fine-grained diagonal search, and Bayesian optimization. For a fair comparison with gradient descent on weights, we aim to find the best performance on the subspace through \textbf{training} data. This separates the question of optimization capacity from train-test generalization and gives the task-vector subspace a favorable comparison. Nonetheless, we still report the best \textbf{test} performance among these methods as the subspace best. For Bayesian optimization (BO), we use $2*T+1$ warmup points and sample $50$ points in total, where $T$ is the number of task vectors, except for TIES and TSV-M on language models, for which we sample $20$ points in total. To the best of our knowledge, we are the first to apply Bayesian Optimization to the coefficients of vision model merging.

For vision models, the results are in \cref{tab:cv_subspace_best}. BO and subspace GD have similar performance. Since Subspace GD is optimized based on cross-entropy loss, there is an accuracy-loss mismatch that makes BO have better accuracy. Nonetheless, Subspace GD always has the lowest training loss. The fine-grained diagonal search and BO have similar performance on TIES, since it only has one coefficient. Since BO and subspace GD have similar performance, we only run subspace GD for DARE, and only run fine-grained grid search for TSV-M.

For language models, the results are in \cref{tab:llm_subspace_best}. BO has higher scores (accuracy) than subspace GD since the accuracy-loss mismatch is more severe in LLMs. We did not do Subspace GD on Qwen3-4B since it requires huge GPU memory that is beyond our computational resources.

\begin{table}[h]
    \centering
    \caption{\textbf{Finding the best average accuracy on the task-vector subspace for vision models.} Bayesian Optimization (BO) and Subspace GD have similar performance.}
    \label{tab:cv_subspace_best}
    \begin{tabular}{llccc}
        \toprule
        \textbf{Merging}
        & \textbf{Method}
        & \textbf{ViT-B-32}
        & \textbf{ViT-B-16}
        & \textbf{ViT-L-14} \\
        \midrule

        \multirow{2}{*}{TA}
        & BO
        & \textbf{65.10}
        & \textbf{70.24}
        & \textbf{79.58} \\
        & Subspace GD
        & 65.07
        & 68.90
        & 79.31 \\

        \midrule

        \multirow{2}{*}{TIES}
        & BO
        & \textbf{67.34}
        & \textbf{72.49}
        & \textbf{77.75} \\
        & Subspace GD
        & 66.72
        & 71.84
        & -- \\

        \bottomrule
    \end{tabular}
\end{table}

\begin{table}[h]
\centering
\caption{\textbf{Finding the best average score on the task-vector subspace for language models.} If Subspace GD is worse than coefficient search, we do grid search instead. Bayesian Optimization (BO) generally has the best performance.}
\label{tab:llm_subspace_best}
\resizebox{\linewidth}{!}{
\begin{tabular}{llcccc}
\toprule
\textbf{Merging}
& \textbf{Method}
& \textbf{Qwen3-0.6B}
& \textbf{Qwen3-1.7B}
& \textbf{Qwen3-4B}
& \textbf{Llama-3.2-1B} \\
\midrule

\multirow{2}{*}{TA}
& BO
& \textbf{70.20} & \textbf{80.51} & \textbf{81.13} & 76.40 \\
& Subspace GD / Grid Search
& 66.84 & 73.25 & -- & \textbf{77.19} \\

\midrule

\multirow{2}{*}{TIES}
& BO
& \textbf{54.09} & 63.50 & \textbf{68.38} & \textbf{57.72} \\
& Subspace GD / Grid Search
& 53.55 & \textbf{63.78} & -- & 57.67 \\

\midrule

\multirow{2}{*}{DARE}
& BO
& \textbf{68.58} & \textbf{77.41} & \textbf{80.04} & 76.57 \\
& Subspace GD / Grid Search
& 68.52 & 73.57 & -- & \textbf{77.52} \\

\midrule

\multirow{2}{*}{TSV-M}
& BO
& \textbf{69.71} & 78.52 & \textbf{79.74} & 79.34 \\
& Subspace GD / Grid Search
& 69.52 & \textbf{78.67} & -- & \textbf{80.48} \\

\bottomrule
\end{tabular}
}
\end{table}

\clearpage
\subsection{Full-weight vs. LoRA fine-tuning and the effect of initialization points} \label{sec:lora}

In this section, we compare the performance of full-weight fine-tuning and LoRA fine-tuning \citep{hu2022lora} for LLM model merging.  As shown in \cref{tab:lll_gd_lora}, LoRA fine-tuning usually has better performance on merging tasks. We believe this is because the auxiliary dataset is small, and LoRA has fewer parameters than the full weights, reducing the risk of overfitting. We report the LoRA fine-tuning results in \cref{fig: main}. We also test two initialization points. 
Consistent with \cref{sec: exp setup}, we report results that initialize from the naive baseline selected between coefficients of $1/T$ and $1.0$ for each model and merging method in \cref{fig: main} and \cref{tab:full_comparison}.

\begin{table}[ht]
\centering
\caption{\textbf{Average score of full-weight fine-tuning and LoRA fine-tuning across different merging methods and initialization points.} We set the rank of LoRA to $16$, and test two initialization points in the task-vector subspace: all coefficients set to the average $1/T$ or $1.0$. We did not evaluate full-weight fine-tuning on Qwen3-4B due to computational resource constraints.
Overall, LoRA fine-tuning has better performance on the merging tasks. }
\label{tab:lll_gd_lora}
\small
\begin{tabular}{llcccc}
\toprule
\textbf{Merging}
& \textbf{Method}
& \textbf{Qwen3-0.6B}
& \textbf{Qwen3-1.7B}
& \textbf{Qwen3-4B}
& \textbf{Llama-3.2-1B} \\
\midrule

\multirow{2}{*}{Pretrained}
& full-weight fine-tuning
& 53.76 & 58.42 & -- & 51.83 \\
& LoRA fine-tuning
& 60.77 & 65.02 & 70.05 & 53.11 \\

\midrule

\multirow{4}{*}{TA}
& full-weight fine-tuning ($1/T$)
& 70.22 & 75.05 & -- & 61.03 \\
& full-weight fine-tuning ($1.0$)
& 70.98 & 76.10 & -- & 78.83 \\
& LoRA fine-tuning ($1/T$)
& 69.94 & 76.93 & 83.91 & 80.77 \\
& LoRA fine-tuning ($1.0$)
& 73.02 & 81.48 & 83.40 & 82.04 \\

\midrule

\multirow{4}{*}{TIES}
& full-weight fine-tuning ($1/T$)
& 56.68 & 62.92 & -- & 60.86 \\
& full-weight fine-tuning ($1.0$)
& 67.72 & 71.87 & -- & 76.38 \\
& LoRA fine-tuning ($1/T$)
& 62.88 & 69.29 & 77.26 & 68.48 \\
& LoRA fine-tuning ($1.0$)
& 69.67 & 77.01 & 82.02 & 79.83 \\

\midrule

\multirow{4}{*}{DARE}
& full-weight fine-tuning ($1/T$)
& 70.68 & 74.84 & -- & 59.89 \\
& full-weight fine-tuning ($1.0$)
& 71.18 & 75.31 & -- & 77.53 \\
& LoRA fine-tuning ($1/T$)
& 70.42 & 77.40 & 84.16 & 82.42 \\
& LoRA fine-tuning ($1.0$)
& 73.19 & 81.03 & 82.99 & 81.26 \\

\midrule

\multirow{4}{*}{TSV-M}
& full-weight fine-tuning ($1/T$)
& 68.48 & 70.27 & -- & 57.75 \\
& full-weight fine-tuning ($1.0$)
& 75.78 & 81.77 & -- & 83.39 \\
& LoRA fine-tuning ($1/T$)
& 68.63 & 73.82 & 80.23 & 76.67 \\
& LoRA fine-tuning ($1.0$)
& 77.78 & 81.85 & 87.23 & 82.99 \\

\bottomrule
\end{tabular}
\end{table}

\clearpage

\subsection{More data scaling results} \label{sec:more data scaling}

\begin{figure}[h]
    \centering
    \includegraphics[width=1\linewidth]{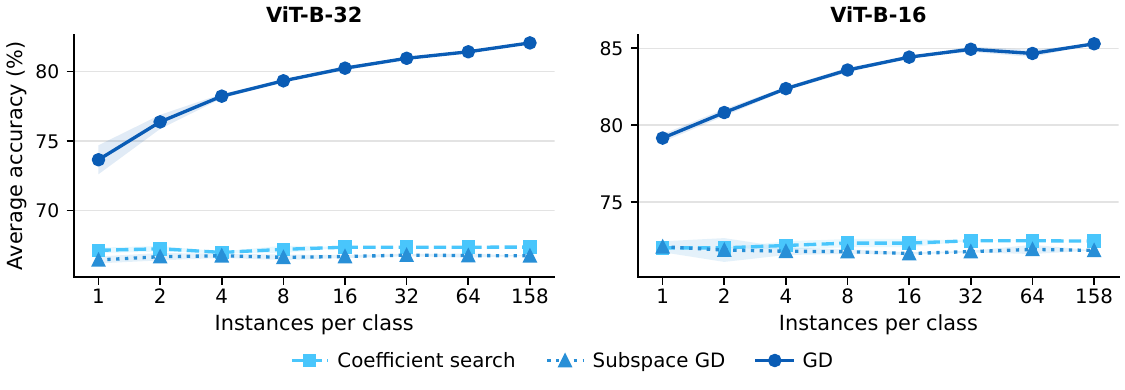}
    \caption{\textbf{Average test accuracy under different sizes and uses of auxiliary data for TIES.} We downsample the auxiliary data and compare three different uses of auxiliary data for TIES. We run three random seeds for downsampling, and report error bars as shaded regions. The findings are consistent with the TA's result in \cref{fig: data scaling ta}.}
    \label{fig: data scaling ties}
\end{figure}

As shown in \cref{tab:vision-splits}, some datasets have fewer validation examples. For classes with insufficient validation samples, we use all available samples without replacement. Therefore, the x-axis in \cref{fig: data scaling ta} indicates $\text{min}(\text{all samples}, \text{required instances per class})$.
\cref{fig: data scaling ties} shows the performance of different strategies for using auxiliary data under different dataset sizes for TIES. Since TIES only has one coefficient, the subspace GD and coefficient search have similar performance. However, doing gradient descent on weights (GD) consistently has better accuracy across different dataset sizes.

\subsection{Main results in Tables}

\cref{fig: main}'s numerical values are shown in \cref{tab:full_comparison}. We also include the results on Llama-3.2-1B, and the findings are consistent with Qwen3 models: gradient descent on model weights outperforms optimizing within the task-vector subspace. \cref{tab:ta_data_scaling} and \cref{tab:ties_data_scaling} report the data-scaling results for TA and TIES, respectively.

\begin{table}[ht]
    \centering
    \caption{\textbf{Numerical values of \cref{fig: main}}. We also provide the Llama-3.2-1B results, and the results are consistent.}
    \label{tab:full_comparison}
    \resizebox{\linewidth}{!}{%
        \begin{tabular}{ll|ccc|cccc}
            \toprule
            & & \multicolumn{3}{c|}{\textbf{Vision}}
              & \multicolumn{4}{c}{\textbf{Language}} \\
            \textbf{Merging Method} & \textbf{Baseline}
              & \textbf{ViT-B-32}
              & \textbf{ViT-B-16}
              & \textbf{ViT-L-14}
              & \textbf{Qwen3-0.6B}
              & \textbf{Qwen3-1.7B}
              & \textbf{Qwen3-4B}
              & \textbf{Llama-3.2-1B} \\
            \midrule

            Pretrained
              & gradient descent
              & 78.54 & 82.51 & 84.98
              & 60.77 & 65.02 & 70.05 & 53.11 \\

            \midrule
            TA
              & naive baseline
              & 62.87 & 67.57 & 74.61
              & 60.13 & 64.58 & 75.49 & 74.87 \\
            & coefficient search
              & 62.41 & 68.06 & 76.68
              & 63.62 & 73.65 & 79.99 & 76.22 \\
            & subspace best
              & 65.10 & 70.24 & 79.58
              & 70.20 & 80.51 & 81.13 & 77.19 \\
            & gradient descent
              & 80.53 & 83.77 & 85.85
              & 69.94 & 81.48 & 83.91 & 80.77 \\

            \midrule
            TIES
              & naive baseline
              & 62.25 & 69.52 & 76.27
              & 49.98 & 57.93 & 68.26 & 57.84 \\
            & coefficient search
              & 67.35 & 72.45 & 77.72
              & 50.09 & 57.93 & 67.87 & 58.22 \\
            & subspace best
              & 67.35 & 72.49 & 77.78
              & 54.09 & 63.78 & 68.38 & 57.72 \\
            & gradient descent
              & 82.06 & 85.28 & 87.27
              & 69.67 & 77.01 & 82.02 & 79.83 \\

            \midrule
            DARE
              & naive baseline
              & 62.73 & 67.63 & 74.61
              & 60.46 & 65.64 & 74.70 & 75.15 \\
            & coefficient search
              & 62.38 & 67.94 & 76.71
              & 63.88 & 72.30 & 79.86 & 77.07 \\
            & subspace best
              & 64.55 & 68.97 & 79.27
              & 68.58 & 77.41 & 80.04 & 77.52 \\
            & gradient descent
              & 80.52 & 83.72 & 84.48
              & 70.42 & 81.03 & 84.16 & 82.42 \\

            \midrule
            TSV-M
              & naive baseline
              & 74.79 & 80.38 & 87.29
              & 68.32 & 77.89 & 81.26 & 70.36 \\
            & coefficient search
              & 78.05 & 81.91 & 87.53
              & 69.43 & 78.37 & 79.85 & 79.77 \\
            & subspace best
              & 78.05 & 81.96 & 87.53
              & 69.71 & 78.67 & 79.74 & 80.48 \\
            & gradient descent
              & 83.88 & 86.37 & 88.26
              & 77.78 & 81.85 & 87.23 & 82.99 \\

            \bottomrule
        \end{tabular}%
    }
\end{table}

\begin{table}[ht]
    \centering
    \caption{\textbf{Numerical values of \cref{fig: data scaling ta}}: data scaling for TA.}
    \label{tab:ta_data_scaling}
    \resizebox{\linewidth}{!}{%
        \begin{tabular}{lcccccccc}
            \toprule
            \textbf{Method / Data size} & \textbf{1} & \textbf{2} & \textbf{4} & \textbf{8} & \textbf{16} & \textbf{32} & \textbf{64} & \textbf{158} \\
            \midrule
            \textbf{ViT-B-32} \\
            \hspace{1em}Coefficient search
                & 62.42 $\pm$ 0.00
                & 62.12 $\pm$ 0.25
                & 62.27 $\pm$ 0.26
                & 62.27 $\pm$ 0.25
                & 62.13 $\pm$ 0.25
                & 62.27 $\pm$ 0.25
                & 62.42 $\pm$ 0.00
                & 62.41 \\
            \hspace{1em}Subspace GD
                & 62.86 $\pm$ 0.99
                & 64.26 $\pm$ 0.59
                & 64.39 $\pm$ 0.62
                & 64.56 $\pm$ 0.31
                & 64.71 $\pm$ 0.20
                & 64.31 $\pm$ 0.42
                & 64.66 $\pm$ 0.29
                & 65.08 \\
            \hspace{1em}GD
                & 68.44 $\pm$ 1.24
                & 72.34 $\pm$ 0.52
                & 75.16 $\pm$ 0.02
                & 76.76 $\pm$ 0.17
                & 77.92 $\pm$ 0.05
                & 78.81 $\pm$ 0.06
                & 79.29 $\pm$ 0.12
                & 80.53 \\
            \midrule
            \textbf{ViT-B-16} \\
            \hspace{1em}Coefficient search
                & 68.07 $\pm$ 0.01
                & 68.07 $\pm$ 0.01
                & 67.75 $\pm$ 0.56
                & 68.07 $\pm$ 0.01
                & 68.07 $\pm$ 0.01
                & 68.07 $\pm$ 0.01
                & 68.07 $\pm$ 0.01
                & 68.07 \\
            \hspace{1em}Subspace GD
                & 67.78 $\pm$ 1.55
                & 68.92 $\pm$ 0.53
                & 68.23 $\pm$ 0.38
                & 68.71 $\pm$ 0.19
                & 68.72 $\pm$ 0.23
                & 68.99 $\pm$ 0.28
                & 68.78 $\pm$ 0.09
                & 68.91 \\
            \hspace{1em}GD
                & 74.70 $\pm$ 0.66
                & 76.45 $\pm$ 0.49
                & 78.51 $\pm$ 0.23
                & 80.17 $\pm$ 0.28
                & 81.41 $\pm$ 0.20
                & 82.27 $\pm$ 0.13
                & 82.78 $\pm$ 0.11
                & 83.78 \\
            \bottomrule
        \end{tabular}%
    }
\end{table}

\begin{table}[ht]
    \centering
    \caption{\textbf{Numerical values of \cref{fig: data scaling ties}}: data scaling for TIES.}
    \label{tab:ties_data_scaling}
    \resizebox{\linewidth}{!}{%
        \begin{tabular}{lcccccccc}
            \toprule
            \textbf{Method / Data size} & \textbf{1} & \textbf{2} & \textbf{4} & \textbf{8} & \textbf{16} & \textbf{32} & \textbf{64} & \textbf{158} \\
            \midrule
            \textbf{ViT-B-32} \\
            \hspace{1em}Coefficient search
                & 67.12 $\pm$ 0.20
                & 67.23 $\pm$ 0.20
                & 66.97 $\pm$ 0.06
                & 67.19 $\pm$ 0.27
                & 67.35 $\pm$ 0.00
                & 67.35 $\pm$ 0.00
                & 67.35 $\pm$ 0.00
                & 67.35 \\
            \hspace{1em}Subspace GD
                & 66.43 $\pm$ 0.25
                & 66.66 $\pm$ 0.26
                & 66.73 $\pm$ 0.15
                & 66.62 $\pm$ 0.19
                & 66.68 $\pm$ 0.11
                & 66.77 $\pm$ 0.08
                & 66.74 $\pm$ 0.14
                & 66.73 \\
            \hspace{1em}GD
                & 73.65 $\pm$ 1.04
                & 76.37 $\pm$ 0.51
                & 78.23 $\pm$ 0.16
                & 79.34 $\pm$ 0.06
                & 80.25 $\pm$ 0.16
                & 80.96 $\pm$ 0.02
                & 81.43 $\pm$ 0.02
                & 82.07 \\
            \midrule
            \textbf{ViT-B-16} \\
            \hspace{1em}Coefficient search
                & 72.01 $\pm$ 0.00
                & 72.01 $\pm$ 0.00
                & 72.17 $\pm$ 0.27
                & 72.32 $\pm$ 0.27
                & 72.32 $\pm$ 0.27
                & 72.48 $\pm$ 0.00
                & 72.48 $\pm$ 0.00
                & 72.45 \\
            \hspace{1em}Subspace GD
                & 72.08 $\pm$ 0.35
                & 71.87 $\pm$ 0.77
                & 71.79 $\pm$ 0.25
                & 71.77 $\pm$ 0.15
                & 71.65 $\pm$ 0.07
                & 71.77 $\pm$ 0.05
                & 71.91 $\pm$ 0.30
                & 71.85 \\
            \hspace{1em}GD
                & 79.15 $\pm$ 0.25
                & 80.81 $\pm$ 0.22
                & 82.37 $\pm$ 0.11
                & 83.58 $\pm$ 0.16
                & 84.42 $\pm$ 0.03
                & 84.93 $\pm$ 0.17
                & 84.66 $\pm$ 0.23
                & 85.29 \\
            \bottomrule
        \end{tabular}%
    }
\end{table}

\newpage
\subsection{S-RandOpt implementation details} \label{sec:srandopt_details}

In \cref{sec:srandopt}, we calculate the gradient from the best coefficient search point for vision models, and from the subspace best for the language models. From this initialization point $\theta_{init}$, we construct a bounded region, which we call the \textit{Sample Space}. Let $d$ denote the component of gradient descent that is orthogonal to the task-vector subspace. We define the Sample Space as
\[
\mathcal{S}=\left\{\theta_{\mathrm{init}}+\sum_{i=1}^{T} a_i \tilde{\tau}_i+b\,d\;:\;
a_i \in [-\alpha,\alpha],\;b \in [0,\beta]\right\}.\]

Thus, each task-vector coefficient is allowed to vary within $[-\alpha,\alpha]$, while the coefficient along the additional descent direction varies within $[0,\beta]$. We set $(\alpha,\beta)=(0.1,0.01)$ for vision models and $(\alpha,\beta)=(0.05,0.0005)$ for language models.

\clearpage

\section{More analyses}

\subsection{Complete analysis of initializations for gradient descent} \label{app: init matters}

\begin{figure}[ht!]
    \centering

    \begin{subfigure}{0.32\textwidth}
        \includegraphics[width=\linewidth]{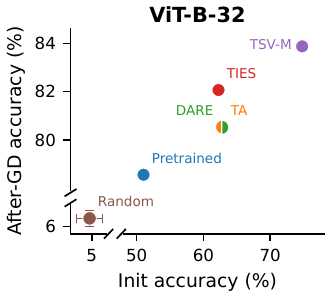}
    \end{subfigure}\hfill
    \begin{subfigure}{0.32\textwidth}
        \includegraphics[width=\linewidth]{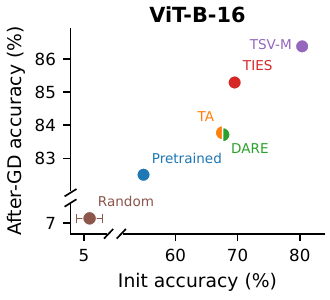}
    \end{subfigure}\hfill
    \begin{subfigure}{0.32\textwidth}
        \includegraphics[width=\linewidth]{figure/init_scatter/vision/9_tasks_dtd-eurosat-fer2013-food101-gtsrb-mnist-resisc45-stanford-cars-sun397__vit-l-14.pdf}
    \end{subfigure}

    \vspace{0.3cm}

    \begin{subfigure}{0.32\textwidth}
        \includegraphics[width=\linewidth]{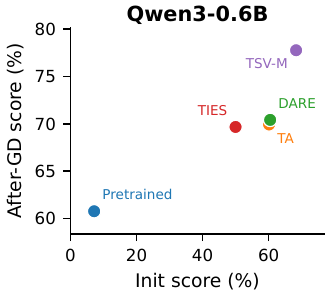}
    \end{subfigure}\hfill
    \begin{subfigure}{0.32\textwidth}
        \includegraphics[width=\linewidth]{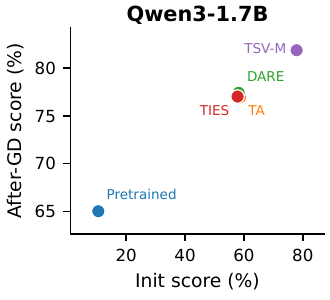}
    \end{subfigure}\hfill
    \begin{subfigure}{0.32\textwidth}
        \includegraphics[width=\linewidth]{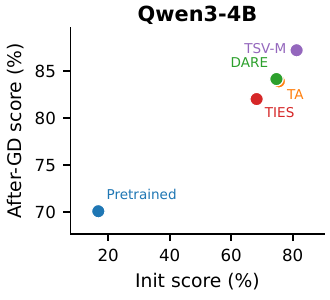}
    \end{subfigure}

    \caption{\textbf{Gradient Descent (GD) Initialization results across vision and language models.} For vision, we measure average test accuracy, and report the average accuracy of three random initialization points. Init accuracy is the naive coefficient’s performance.}
    \label{fig:app_init_scatter}
\end{figure}

\cref{fig:app_init_scatter} shows the full results of gradient descent with different initializations. Across all models, the message is consistent: merging methods provide strong initializations for gradient descent, and they are better than fine-tuning from the pretrained model or random initializations.

\clearpage

\subsection{2-task setting} \label{sec:2-task-scenario}

For the two-task scenario, we evaluate two task combinations: EuroSAT \citep{helber2019eurosat} and GTSRB \citep{stallkamp2011german}, as well as MNIST \citep{lecun1998mnist} and FER2013 \citep{goodfellow2013challenges}. We tested both regularized and unregularized versions of gradient descent. We apply the $\ell_2$ norm from the starting point, L2-SP, as the regularization added to the loss function:
$\frac{\alpha}{2} \|\theta - \theta_{init}\|^2_2$. $\theta_{init}$ is the initialization point for gradient descent, and we set $\alpha =3$.

\begin{figure}[ht!]
    \centering

    \begin{subfigure}{\textwidth}
        \includegraphics[width=\linewidth]{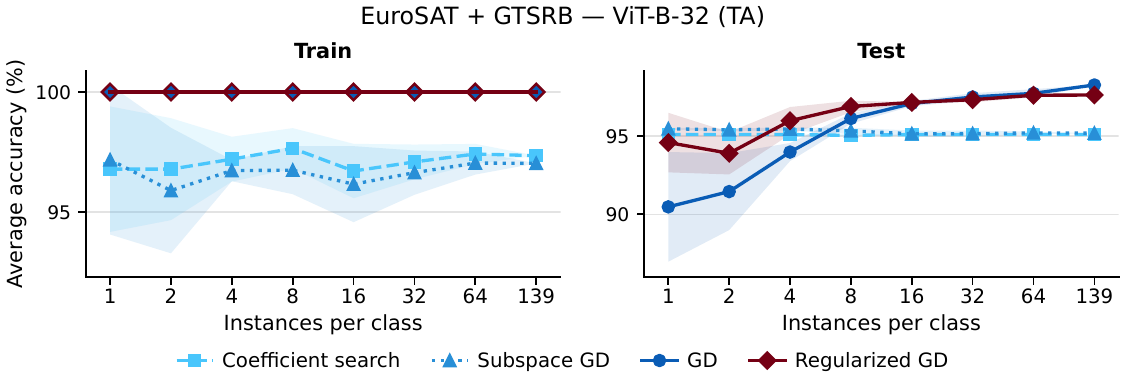}
    \end{subfigure}

    \vspace{0.3cm}

    \begin{subfigure}{\textwidth}
        \includegraphics[width=\linewidth]{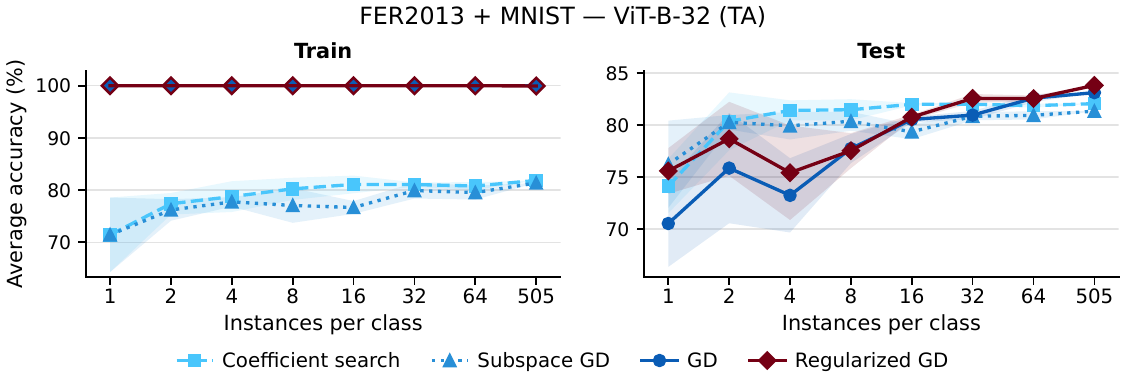}
    \end{subfigure}\hfill

    \caption{\textbf{Train and test set accuracy of different strategies in 2-task scenarios.} GD overfits when data are scarce, and additional regularization improves performance. On the other hand, the test accuracy of coefficient search and subspace GD is largely insensitive to data size.}
    \label{fig:2_task_train_test}
\end{figure}

\cref{fig:2_task_train_test} shows the experimental results. GD and regularized GD learn the training set perfectly across all data sizes in two scenarios, while coefficient search and subspace GD struggle even with more data. Overall, when data are scarce, GD may lead to overfitting, but the risk is smaller when data are sufficient. In contrast, coefficient search and subspace GD do not fit the training set well, so their test performance changes little as more data become available. The appropriate regularization strength for balancing the bias-variance trade-off is a key factor in determining how auxiliary data should be used.

\subsubsection{Ablation Study: Task diversity}

While the 9-task setting in \cref{fig: data scaling ta} shows that gradient descent outperforms coefficient optimization across all downsampling scenarios, the 2-task setting in \cref{fig:2_task_train_test} yields a different conclusion. The two settings differ in both dataset size and task diversity. In this section, we control the dataset size across the two settings, and investigate whether task diversity determines which optimization strategy performs better. 

Since we control the dataset size by fixing the number of instances per class, we control the number of classes to keep the dataset size the same across the 2-task and 9-task settings. We downsample the number of classes in the 9-task setting to match the 2-task setting. Specifically, we keep the number of classes approximately the same across 9 tasks. In \cref{fig:2_task_train_test}, EuroSAT + GTSRB has 53 classes, and FER2013 + MNIST has 17 classes. We test 3 random seeds for class selection. If having more tasks makes merging easier, then we will see that GD remains consistently better even after downsampling classes.

\begin{figure}[ht!]
    \centering

    \begin{subfigure}{0.32\textwidth}
        \includegraphics[width=\linewidth]{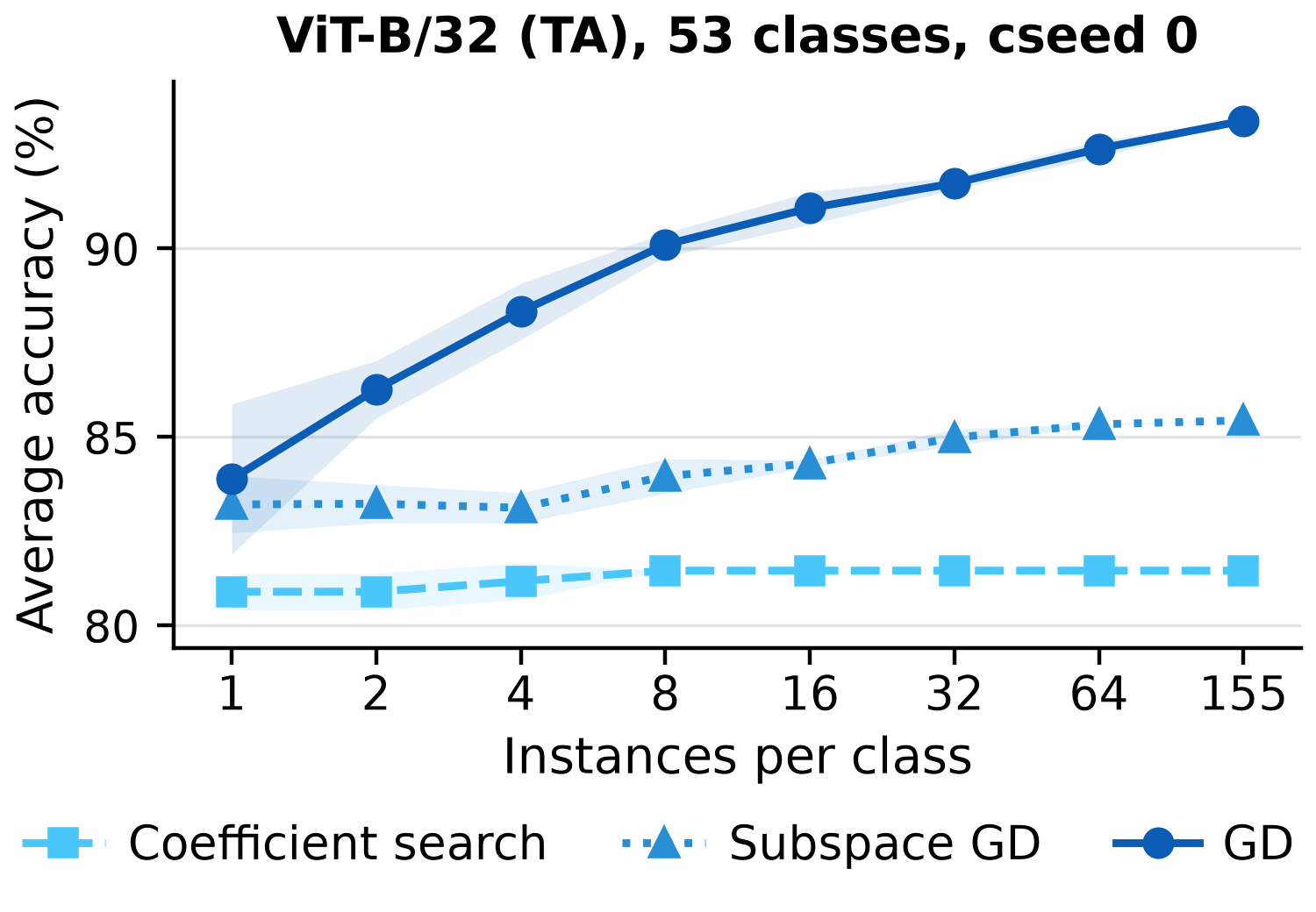}
    \end{subfigure}\hfill
    \begin{subfigure}{0.32\textwidth}
        \includegraphics[width=\linewidth]{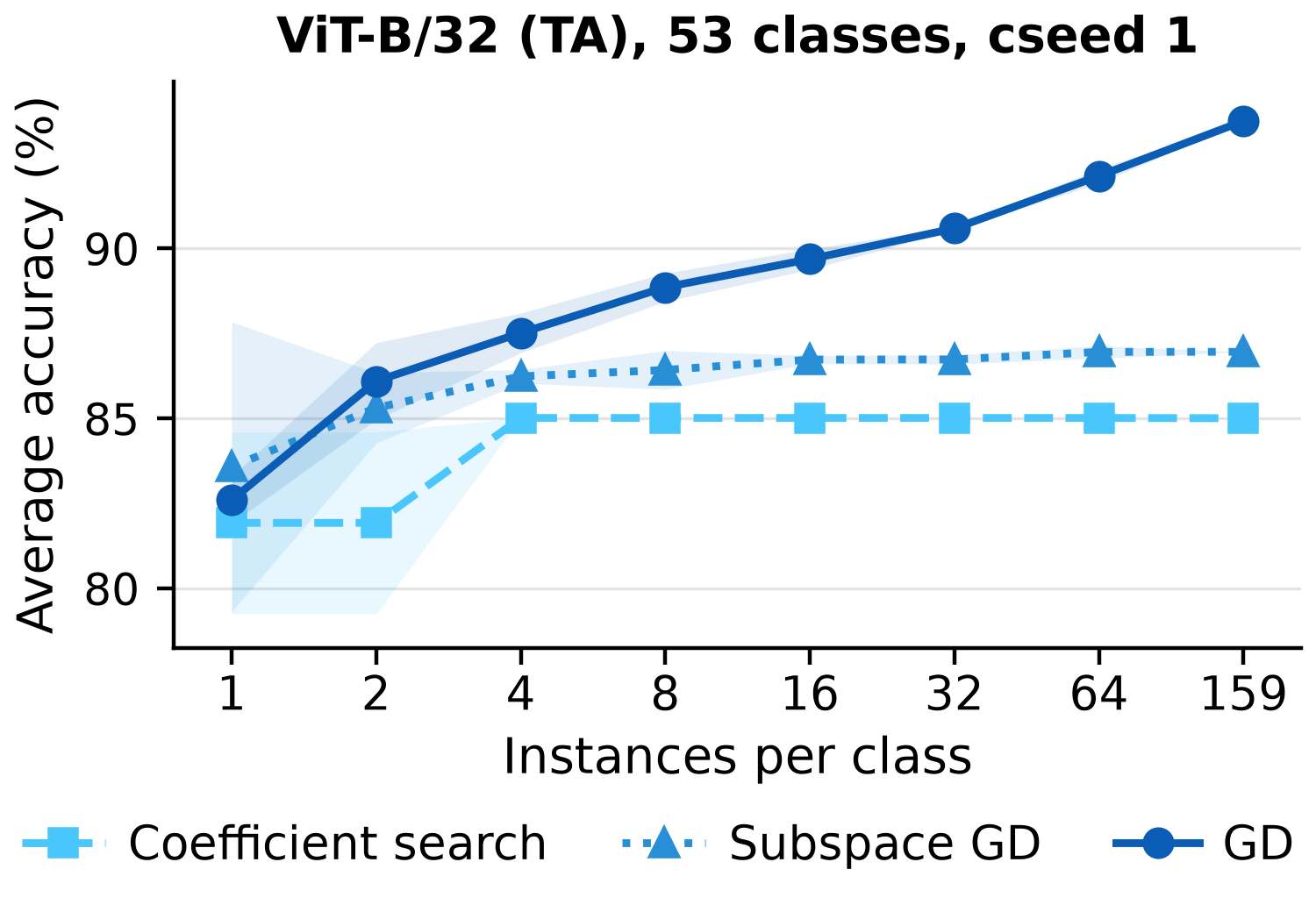}
    \end{subfigure}\hfill
    \begin{subfigure}{0.32\textwidth}
        \includegraphics[width=\linewidth]{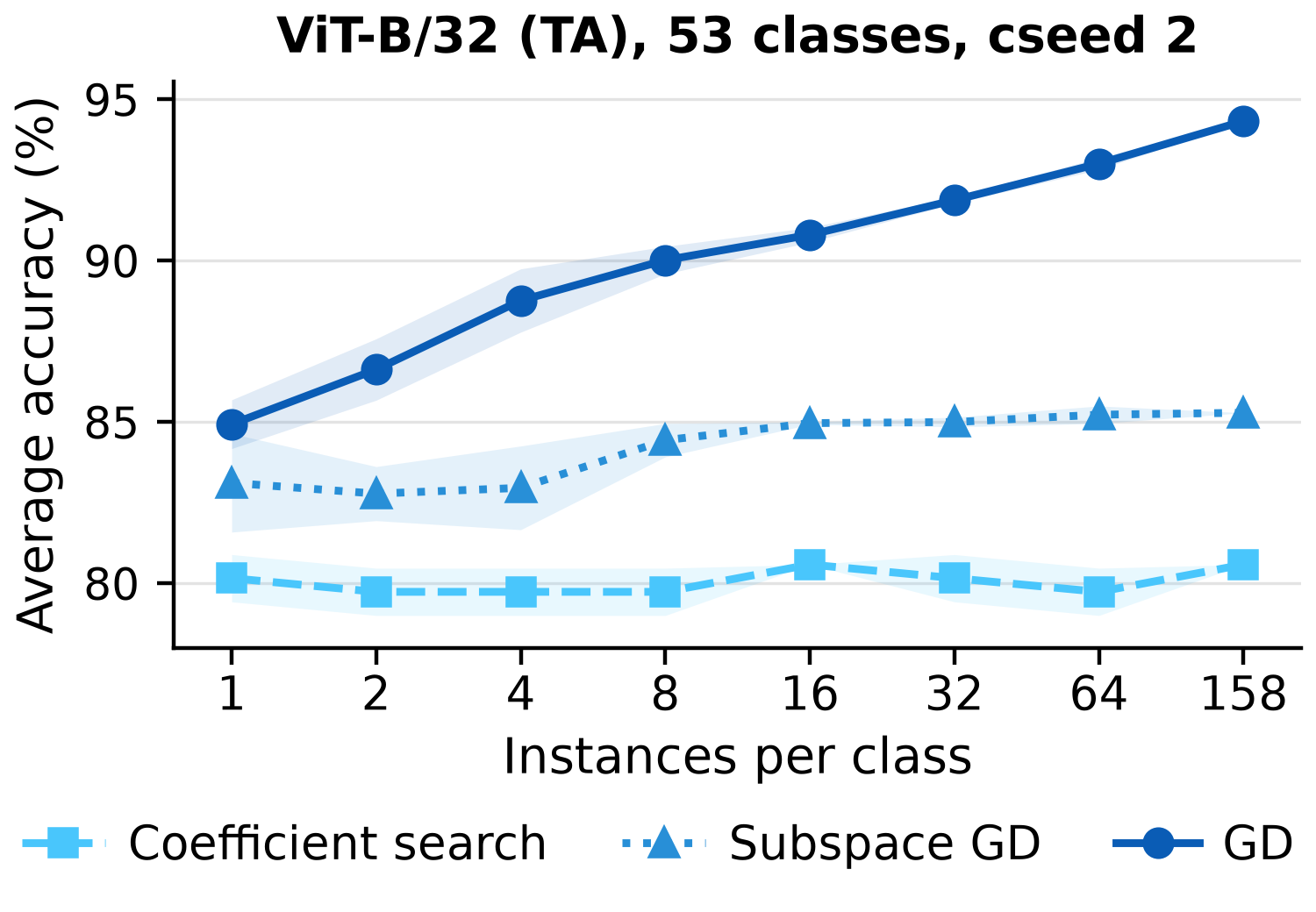}
    \end{subfigure}

    \vspace{0.3cm}

    \begin{subfigure}{0.32\textwidth}
        \includegraphics[width=\linewidth]{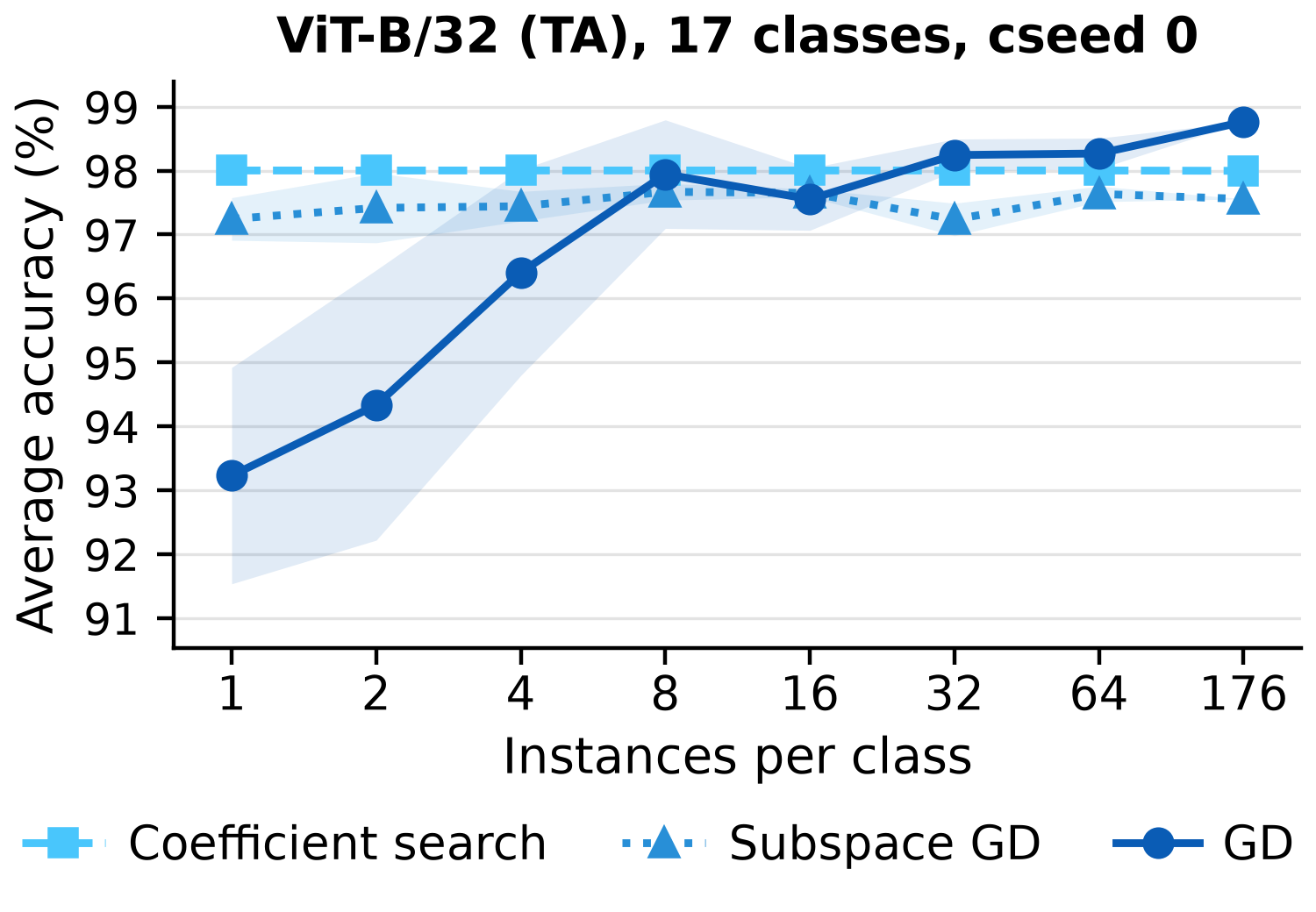}
    \end{subfigure}\hfill
    \begin{subfigure}{0.32\textwidth}
        \includegraphics[width=\linewidth]{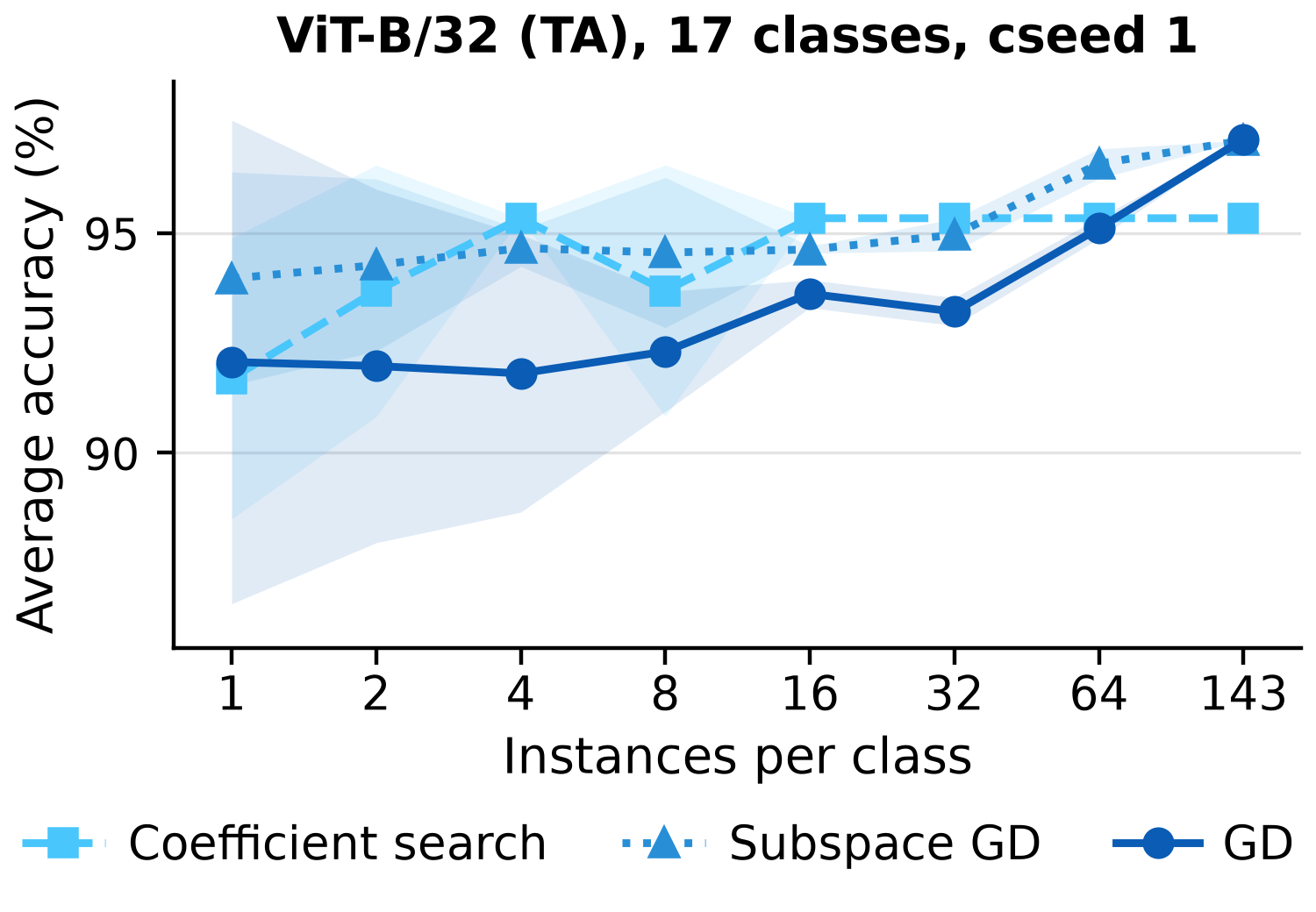}
    \end{subfigure}\hfill
    \begin{subfigure}{0.32\textwidth}
        \includegraphics[width=\linewidth]{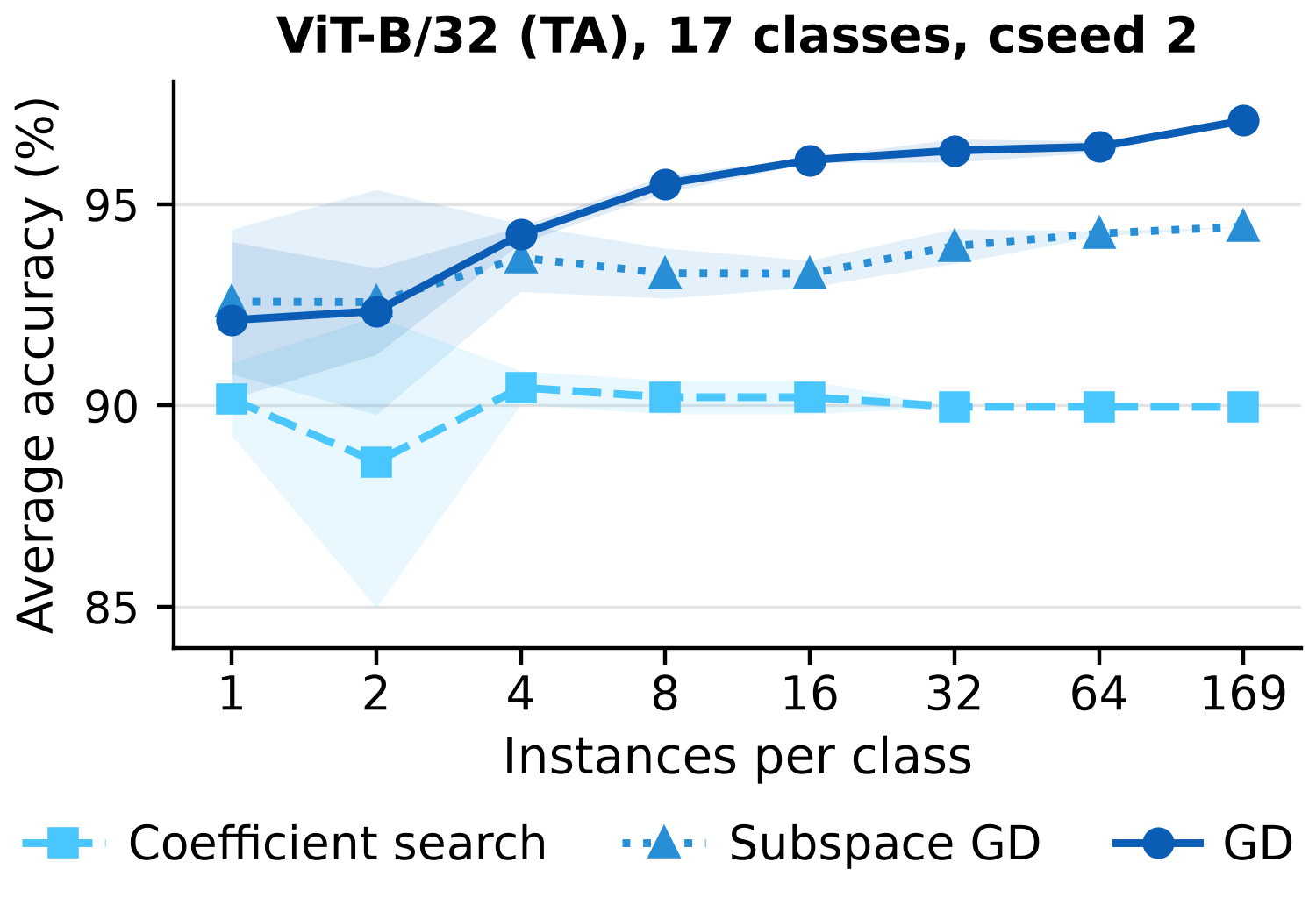}
    \end{subfigure}

    \caption{\textbf{Downsampling classes in the 9-task setting for TA on ViT-B-32.} We test 3 class selection seeds (cseed) for downsampling classes. The first row matches dataset size of EuroSAT + GTSRB in \cref{fig:2_task_train_test}, and the second row matches FER2013 + MNIST. The trends are similar to \cref{fig:2_task_train_test}.}
    \label{fig:smaller_9_tasks}
\end{figure}

\cref{fig:smaller_9_tasks} shows the results. In the full-class 9-task setting in \cref{fig: data scaling ta}, GD consistently outperforms coefficient optimization regardless of instance selection. However, when the number of classes is downsampled, the two methods perform similarly, and GD can even perform worse. The results indicate that dataset size, rather than task diversity, is the main factor determining which strategy best leverages the auxiliary data.

\clearpage

\subsection{Bias-variance tradeoff} \label{app: bv_tradeoff}
\paragraph{Cross-entropy bias-variance decomposition.}
Following the prior works \citep{heskes1998bias,yang2020rethinking}, we decompose the cross-entropy test loss into bias and variance.
Let $D$ denote a randomly sampled training set, $C$ the number of classes,
and $y$ the ground-truth label of a fixed test input $x$.
Let $q_c = \textbf{1}[c=y]$ be the one-hot label and
$p_{D,c}$ be the predicted probability of class $c$ from the model trained on $D$.
The cross-entropy loss is
\begin{equation}
H(q,p_D) = -\sum_{c=1}^{C} q_c \log p_{D,c}
= -\log p_{D,y}.
\end{equation}
Taking the expectation over training sets, we obtain
\begin{equation}
\mathbb{E}_D[H(q,p_D)] = \mathbb{E}_D[-\log p_{D,y}]
= \underbrace{-\log \tilde p_y}_{\text{Bias}}
+ \underbrace{\mathbb{E}_D\left[
\mathrm{KL}(\tilde p \,\|\, p_D)
\right]}_{\text{Variance}}.
\end{equation}
Here, $\tilde p_c =
\frac{\exp(\mathbb{E}_D[\log p_{D,c}])}
{\sum_{j=1}^{C}\exp(\mathbb{E}_D[\log p_{D,j}])}$
is the central prediction, and
$\mathrm{KL}(\tilde p \| p_D)
= \sum_{c=1}^{C}\tilde p_c \log(\tilde p_c/p_{D,c})$
measures the difference between the central prediction and
the prediction from each training set.

\paragraph{Empirically measuring the bias-variance decomposition.}
For each training data size, we draw $20$ different training sets using random seeds. For plotting the confidence intervals, we use $200$ bootstrap samples.

\begin{figure}[h!]
    \centering
\includegraphics[width=\linewidth]{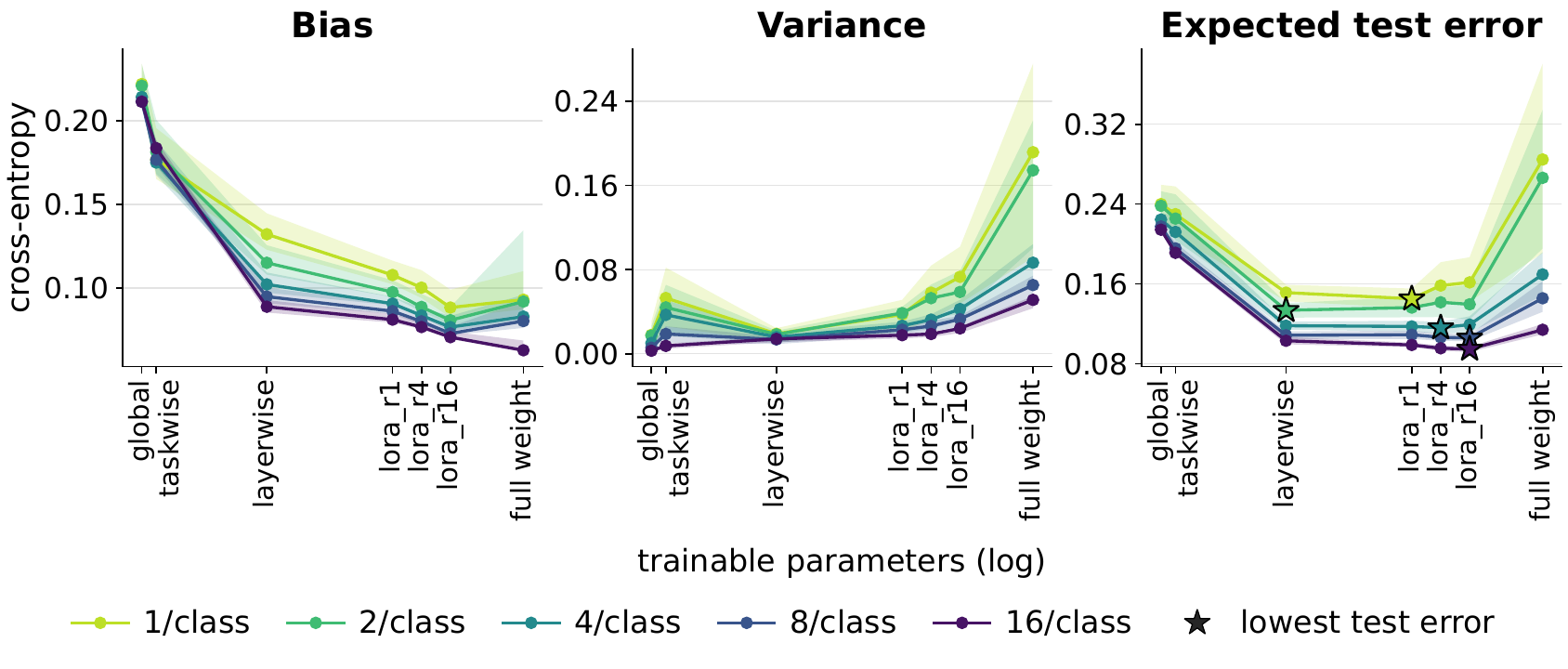}
    \caption{\textbf{Bias-variance decomposition of merging EuroSAT and GTSRB on VIT-B-32}. We use the trainable parameters as a proxy for model complexity, and control the instances per class to vary the training data size. The test set is fixed. We test three types of coefficient optimization (global, taskwise, and layerwise merging), LoRA with three ranks ($r=1,4,16$), and full-weight fine-tuning. For each training data size, a star indicates the lowest expected test error. The results are similar to merging EuroSAT and MNIST in \cref{fig: bv_tradeoff_eurosat_mnist}.}
    \label{fig: bv_tradeoff_eurosat_gtsrb}
\end{figure}

\paragraph{More experimental results.} The results on merging EuroSAT and GTSRB are in \cref{fig: bv_tradeoff_eurosat_gtsrb}. The trends are similar to those when merging EuroSAT and MNIST, as shown in \cref{fig: bv_tradeoff_eurosat_mnist}. But as expected, the optimal model complexity differs, indicating that each scenario requires different regularization strength.

\clearpage

\subsection{Test-time adaptation full results} \label{sec:tta_appendix}

\paragraph{Experiment Setup.} 

We compare GD with two test-time adaptation methods for model merging, AdaMerging \citep{yang2024adamerging} and DivMerge \citep{touayouch2026divmerge}. Both methods learn the merging coefficients on unlabeled test data: AdaMerging minimizes the prediction entropy of the merged model, while DivMerge minimizes the Jensen--Shannon divergence between the predictions of the merged model and those of the task-specific models. We apply both task-wise and layer-wise merging, which are described in detail in \cref{sec:detailed_related_works}, and initialize the merging coefficients to $1/T$. 

For GD, we use the predictions of the task-specific models on the unlabeled test data as supervision. We consider two types of labels: soft labels use the softmax outputs of the task-specific models directly, while hard labels convert them into one-hot labels. All methods use 1,000 unlabeled test samples per task.

\paragraph{Experiment results.} 

\cref{tab:tta_appendix} reports the full experiment results. GD on weights has the best performance in smaller models. Layer-wise DivMerge has more trainable parameters than the task-wise version, yet fewer trainable parameters than full-weight fine-tuning. Therefore, it has less risk of overfitting the unlabeled test data, which turns out to achieve the best performance in ViT-L-14. We expect applying regularization on GD will improve its performance, as shown in \cref{sec: extreme case}. Overall, the empirical results demonstrate that the bias-variance tradeoff is the key principle in deciding how strong the regularization should be when using auxiliary data.

\begin{table}[h]
    \centering
    \caption{\textbf{Full comparison results of test-time adaptation methods for TA.} }
    \label{tab:tta_appendix}
    \begin{tabular}{lccc}
        \toprule
        Method & ViT-B-32 & ViT-B-16 & ViT-L-14 \\
        \midrule
        (task-wise) AdaMerging  & 62.21 & 66.80 & 76.63 \\
        (task-wise) DivMerge    & 65.43 & 69.62 & 78.49 \\
        (layer-wise) AdaMerging & 75.67 & 79.67 & 87.42 \\
        (layer-wise) DivMerge   & 80.56 & 84.48 & \textbf{89.90} \\
        \midrule
        GD Soft
            & \textbf{82.43} & \textbf{85.51} & 88.15 \\
        GD Hard
            & 82.08 & 85.12 & 88.33 \\
        \bottomrule
    \end{tabular}
\end{table}

\clearpage

\subsection{LLM performance on control tasks}
When merging LLMs, the merged model should not only acquire the capabilities of the expert models, but also retain the general capabilities of the pre-trained model. For example, a multilingual medical assistant merged from a multilingual expert and a medical expert should still retain capabilities such as mathematical reasoning and code generation. Since gradient descent optimizes the model weights on the target tasks only, it may cause forgetting of such capabilities \citep{luo2025empirical}. We therefore compare it with merging within the task-vector subspace on control tasks, i.e., tasks not targeted by any expert model. We use GSM8K \citep{cobbe2021training} and MBPP \citep{austin2021program} as control tasks, which evaluate mathematical reasoning and code generation, respectively. 

\cref{tab:control_tasks} reports the results on Qwen3 models. Overall, gradient descent does not systematically cause more forgetting than coefficient search. On the Qwen3-1.7B model, it improves the average accuracy on both control tasks (GSM8K: 64.8 $\rightarrow$ 74.5; MBPP: 41.7 $\rightarrow$ 49.8), reaching a level comparable to the pre-trained model. On the Qwen3-0.6B model, it improves GSM8K (40.7 $\rightarrow$ 50.6) but degrades MBPP (30.7 $\rightarrow$ 25.7). Moreover, coefficient search itself can cause severe forgetting: on the Qwen3-1.7B model, TA drops to 21.7 on MBPP, whereas gradient descent from the same merged model reaches 50.0. Staying within the task-vector subspace thus does not guarantee the retention of general capabilities.

\begin{table}[h]
\centering
\caption{\textbf{Performance on target and control tasks for merged LLMs.}
We compare coefficient search within the task-vector subspace (Coeff.) with gradient descent on LoRA parameters (GD).
Seen denotes the average performance on the target tasks; GSM8K and MBPP are control tasks not targeted by any expert model.
All results are evaluated on 300 samples per task.
$^\dagger$The pre-trained model is evaluated with raw few-shot prompts instead of the chat template and serves only as a reference.
Avg.\ denotes the average over the four merging methods, excluding the pre-trained model. The better result of each pair is shown in bold.}
\label{tab:control_tasks}
\begin{tabular}{llcccccc}
\toprule
& & \multicolumn{2}{c}{\textbf{Seen}} & \multicolumn{2}{c}{\textbf{GSM8K}} & \multicolumn{2}{c}{\textbf{MBPP}} \\
\cmidrule(lr){3-4} \cmidrule(lr){5-6} \cmidrule(lr){7-8}
\textbf{Model} & \textbf{Method} & Coeff. & GD & Coeff. & GD & Coeff. & GD \\
\midrule
\multirow{6}{*}{Qwen3-0.6B}
& Pre-trained$^\dagger$ & \multicolumn{2}{c}{--} & \multicolumn{2}{c}{48.0} & \multicolumn{2}{c}{32.0} \\
& TA    & 63.6 & \textbf{69.9} & 42.0 & \textbf{55.7} & \textbf{32.3} & 26.7 \\
& DARE  & 63.9 & \textbf{70.4} & 32.3 & \textbf{52.0} & \textbf{31.7} & 25.7 \\
& TIES  & 50.1 & \textbf{69.7} & \textbf{58.3} & 49.7 & 30.0 & \textbf{32.7} \\
& TSV-M & 69.4 & \textbf{77.8} & 30.3 & \textbf{45.0} & \textbf{28.7} & 17.7 \\
\cmidrule(l){2-8}
& Avg.  & 61.8 & \textbf{72.0} & 40.7 & \textbf{50.6} & \textbf{30.7} & 25.7 \\
\midrule
\multirow{6}{*}{Qwen3-1.7B}
& Pre-trained$^\dagger$ & \multicolumn{2}{c}{--} & \multicolumn{2}{c}{73.7} & \multicolumn{2}{c}{50.7} \\
& TA    & 73.6 & \textbf{76.9} & 61.3 & \textbf{77.0} & 21.7 & \textbf{50.0} \\
& DARE  & 72.3 & \textbf{77.4} & 72.7 & 72.7 & 44.3 & \textbf{51.3} \\
& TIES  & 57.9 & \textbf{77.0} & 55.0 & \textbf{78.0} & \textbf{51.3} & 48.7 \\
& TSV-M & 78.4 & \textbf{81.8} & 70.0 & \textbf{70.3} & 49.3 & 49.3 \\
\cmidrule(l){2-8}
& Avg.  & 70.6 & \textbf{78.3} & 64.8 & \textbf{74.5} & 41.7 & \textbf{49.8} \\
\bottomrule
\end{tabular}
\end{table}

\end{document}